%% file: map-writeup.tex
\documentclass{article}

\input{zachs_macros}

\usepackage{fancyhdr}
\usepackage{authblk}

\usepackage[top=1in, bottom=1in, left=1in, right=1in]{geometry}
\graphicspath{{.}}

\usepackage{cite}

\author[1]{{del Rosario}, Zachary}
\author[2]{Rupp, Matthias}
\author[2]{Kim, Yoolhee}
\author[2]{Antono, Erin}
\author[2]{Ling, Julia}
\affil[1]{Stanford University, Department of Aeronautics and Astronautics}
\affil[2]{Citrine Informatics}
\date{}
\title{Assessing the Frontier: Active Learning, Model Accuracy, and Multi-objective Materials Discovery and Optimization}

\begin{document}
\maketitle\blfootnote{Corresponding author: Zachary del Rosario, zdr@stanford.edu}


\begin{abstract}
  Discovering novel materials can be greatly accelerated by iterative machine learning-informed proposal of candidates---active learning. However, standard \emph{global error} metrics for model quality are not predictive of discovery performance, and can be misleading. We introduce the notion of \emph{Pareto shell error} to help judge the suitability of a model for proposing material candidates. Further, through synthetic cases and a thermoelectric dataset, we probe the relation between acquisition function fidelity and active learning performance. Results suggest novel diagnostic tools, as well as new insights for acquisition function design.
\end{abstract}

\section{Introduction} \label{sec:intro}

Accelerated design, optimization, and tuning of materials via machine learning is receiving increasing interest in science and industry. A major driver of this interest is the potential to reduce the substantial cost and effort involved in manual development, synthesis, and characterization of large numbers of candidate materials. The primary aim is to reduce the number of both failed candidates and development cycles.

A data-driven approach to achieve this acceleration is \emph{active learning} (AL) \cite{s2012b}, an iterative procedure in which a machine-learning model suggests candidate materials, a selection of which are synthesized, characterized, and fed back into the model to complete a learning iteration. The objective of this procedure varies; in materials informatics it is often to identify promising material candidates by optimizing properties of interest. Recent work has leveraged AL for materials when there is a \emph{single objective} \cite{wang2015nested,aggarwal2016information,urhmt2016q,xue2017informatics,ling2017high,tu2018q,gphs2019q,tgcr2019q,yskmto2018q,p2017q,ph2017q,jsfhts2017q,urhmt2016q,sthtct2015q,nknjty2019q,tbdqda2018q,gmj2019q}. However, new issues arise when optimizing multiple objectives simultaneously, as is frequently the case beyond proof-of-principle settings.

\begin{figure}[!ht]
  \centering
  \begin{minipage}{0.45\textwidth}
    \includegraphics[width=0.95\textwidth]{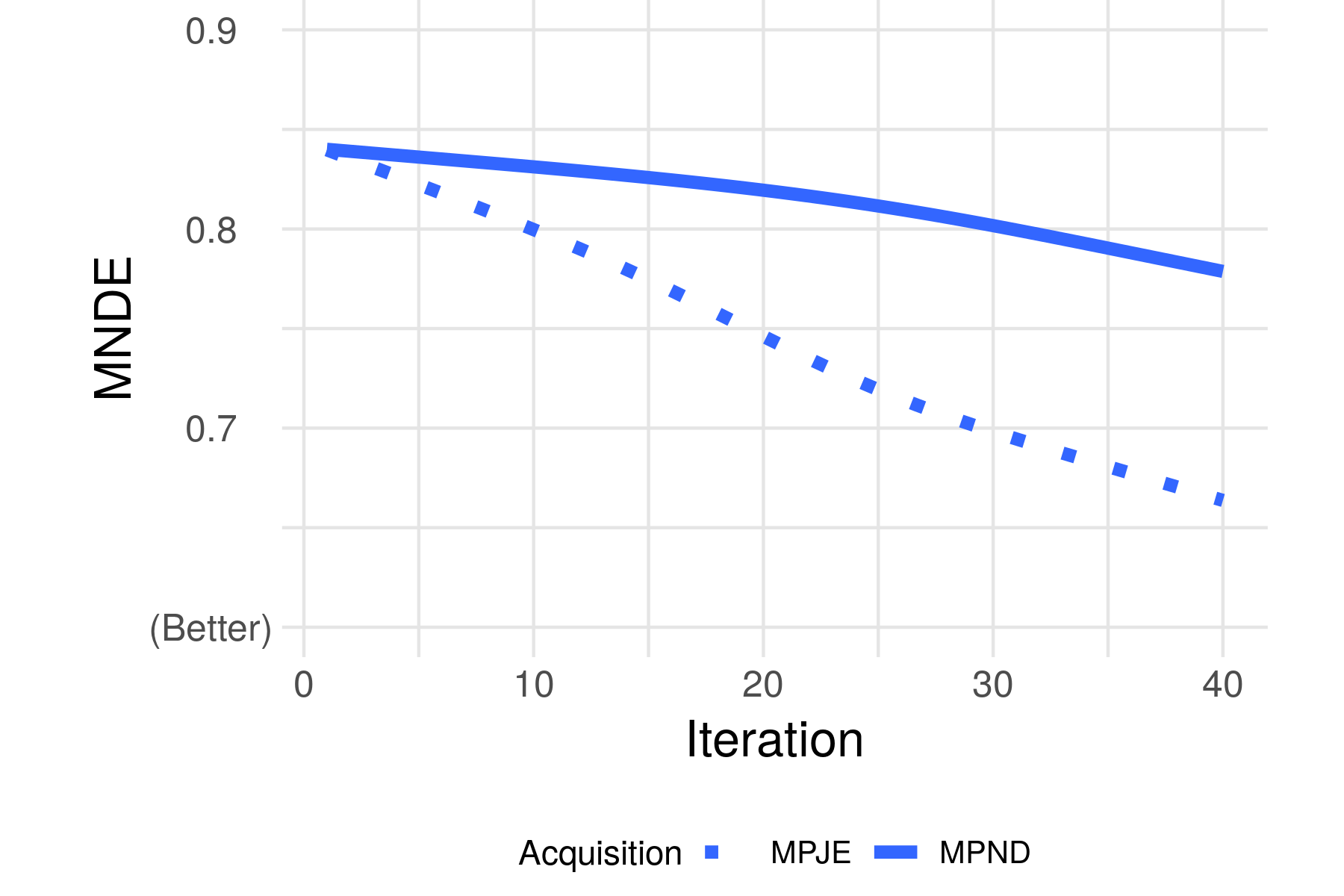}
  \end{minipage} %
  \begin{minipage}{0.45\textwidth}
    \includegraphics[width=0.95\textwidth]{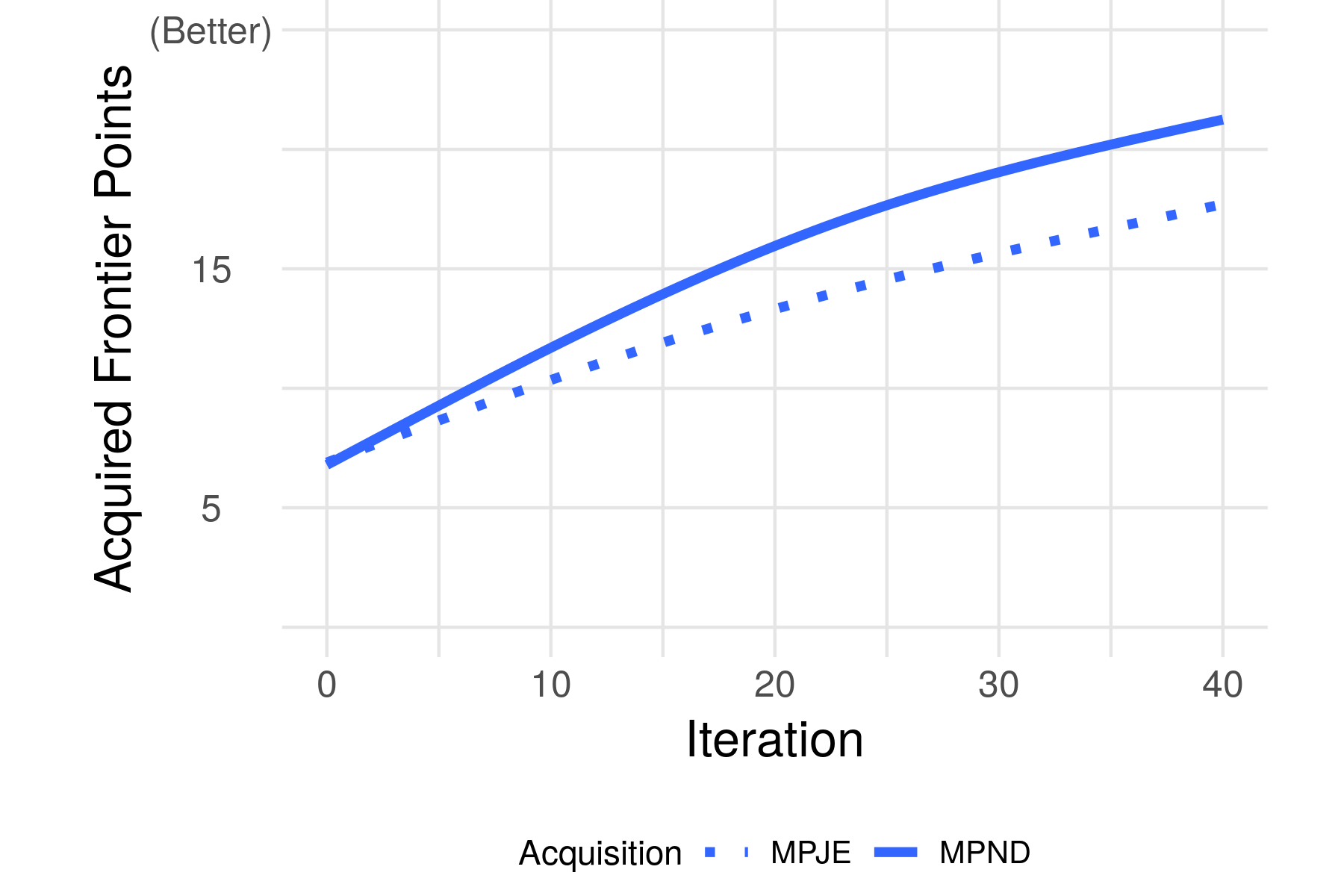}
  \end{minipage} %
  \caption{\emph{Best global error does not guarantee optimal candidate discovery.} Shown are global prediction errors (MNDE, left) and number of optimal candidates found (right). The MPJE decision criterion (solid line) has lower errors, but leads to fewer optimal candidates than the MPND criterion (dashed line). {\small Simulated active learning on a thermoelectric dataset. Estimated mean performance over repetitions (Sec. \ref{sec:results}). MNDE = Mean Non-Dimensional Error \eqref{def:mnde}; MPJE = Maximum Probability of Joint Exceedance \eqref{eq:def-pje}; MPND = Maximum Probability Non-Dominated \eqref{eq:def-pnd}}
  }
  \label{fig:motivating-example}
\end{figure}

Furthermore, the aim to identify promising material candidates is distinct from the aim to improve model accuracy, a frequent objective in AL \cite{settles2009active}. In fact, model accuracy can be at odds with acquiring optimal candidates: Figure \ref{fig:motivating-example} demonstrates that the usual global notion of model accuracy is not necessarily associated with optimal materials discovery. In this work, we introduce a notion of model accuracy more closely associated with rapidly discovering new materials.

We offer two primary contributions: First, we introduce novel concepts to judge the performance of multi-objective AL, with an aim towards the specific concerns of materials discovery. In addition to the usual notion of multi-objective optimality \cite{kochenderfer2019algorithms}---non-dominance---we use criteria informed by the concept of \emph{strata}---recursive non-dominance---from the database literature \cite{roocks2016rPref}. We also introduce \emph{scoped} error metrics, which emphasize regions of interest in performance (property) space, particularly bands about the Pareto frontier we call \emph{Pareto shells}. We demonstrate that the usual global error provides less usable signal for candidate discovery performance, while Pareto shell error---under certain conditions---correctly signals when AL will likely identify performant candidates. Second, we compare multi-objective acquisition functions (also known as \emph{improvement criteria}) in terms of their quantitative and qualitative performance. Specifically, we compare a collection of exploitation/exploration-navigating acquisition functions: Probability of Joint Exceedance (PJE), Hyperplane Probability of Improvement (HPI), and Probability Non-Dominated (PND). These acquisition functions differ in terms of the fidelity with which they represent the Pareto frontier, allowing us to study how this affects AL performance.

In this work we review AL for single-objective materials discovery and relevant concepts from multi-objective optimization, and synthesize materials-relevant concepts for judging AL performance. We introduce a family of acquisition functions (decision-making strategies), and compare their performance across several synthetic datasets and a real thermoelectric dataset. Our results illustrate how new concepts can help articulate the difference between AL strategies, and how model accuracy does---and does not---relate to AL performance in discovering novel materials.

\section{Active Learning for Single-objective Materials Discovery} \label{sec:single-ob}

AL is a specialized problem setting in machine learning related to optimal experimental design \cite{s2012b,m2017c,tbdqda2018q}.
In the materials science context, consider a description~$\vx$ of a material with corresponding observed property of interest $y$, thought to be linked by an unknown function $f: \vx \mapsto y$ which is expensive to evaluate, for example, synthesizing and characterizing a material.
To systematically identify novel materials~$\vx$ with desirable changes in~$y$, a statistical (machine-learning) model $\hat{f}$ is built that predicts the unknown function~$f$.
Trained on an initial set of characterized materials $\tilde{\cX}_0 = \{ \vx_1, \ldots, \vx_{n_0} \}$ with measured properties $\tilde{\cY}_0 = \{ y_1, \ldots, y_{n_0} \}$, the initial model $\hat{f}_0$ is used to select new candidate materials~$\vx_{n_0+1},\ldots,\vx_{n_0+n_1}$.
These are characterized and added to the dataset, $\tilde{\cX}_{1} = \{ \vx_{1},\ldots,\vx_{n_0+n_1} \}$, which is then used to train a new improved model $\hat{f}_1$.
This results in a sequence of datasets $\tilde{\cX}_{k}, \tilde{\cY}_{k}$ for $k = 0, \dots, K$.
While this cycle can in principle be fully automated, new candidates are usually selected by domain experts based on AL suggestions (``human-in-the-loop'').

Various objectives can drive the design of AL strategies; for instance, a common objective in general machine learning is to improve the model~$\hat{f}$.
In materials informatics, the objective is often to identify an improved material~$\vx$ using as few physical experiments as possible \cite{ling2017high}.
We call the design space of experimentally accessible, that is, synthesizable and measurable materials the \emph{global scope}, and call error computed on this scope \emph{global error}.

For a \emph{single objective}, measuring the relative performance of candidates is straightforward (Fig.~\ref{fig:ranking}),
as they can be ranked unambiguously based on their scalar performance, for example, strength-to-weight ratio for structural alloys.
One way to measure the performance of AL is to retrospectively simulate AL on a known dataset and count the number of AL iterations necessary to reach a known top candidate \cite{ling2017high}.
A key decision in designing an AL method is the choice of \emph{acquisition function}.
We will discuss these in greater detail in Section~\ref{sec:multi-ob}; briefly, an acquisition function is a decision rule used to rank potential candidates in an AL context.
Such criteria navigate the ``exploration-exploitation'' trade-off \cite{jljh2018q}:
The algorithm should seek improved candidates, but should also try ``risky'' candidates to improve its model and enable later discoveries \cite{settles2009active}.
Many improvement criteria for the single-objective case have been proposed and tested in the literature \cite{wang2015nested,aggarwal2016information,urhmt2016q,xue2017informatics,ling2017high,seko2014machine}.
The multi-objective setting, however, requires a more nuanced understanding of candidate ranking.

\section{Methods} \label{sec:multi-ob}

\begin{figure}
  \begin{minipage}{0.45\textwidth}
    \includegraphics[width=0.95\textwidth]{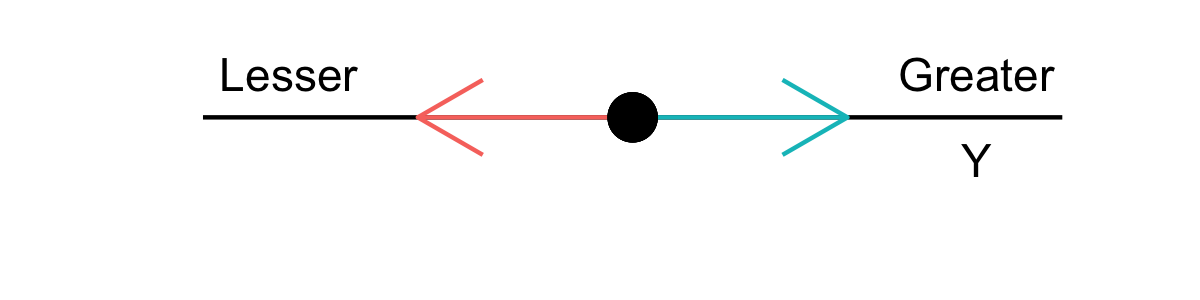}
    \includegraphics[width=0.95\textwidth]{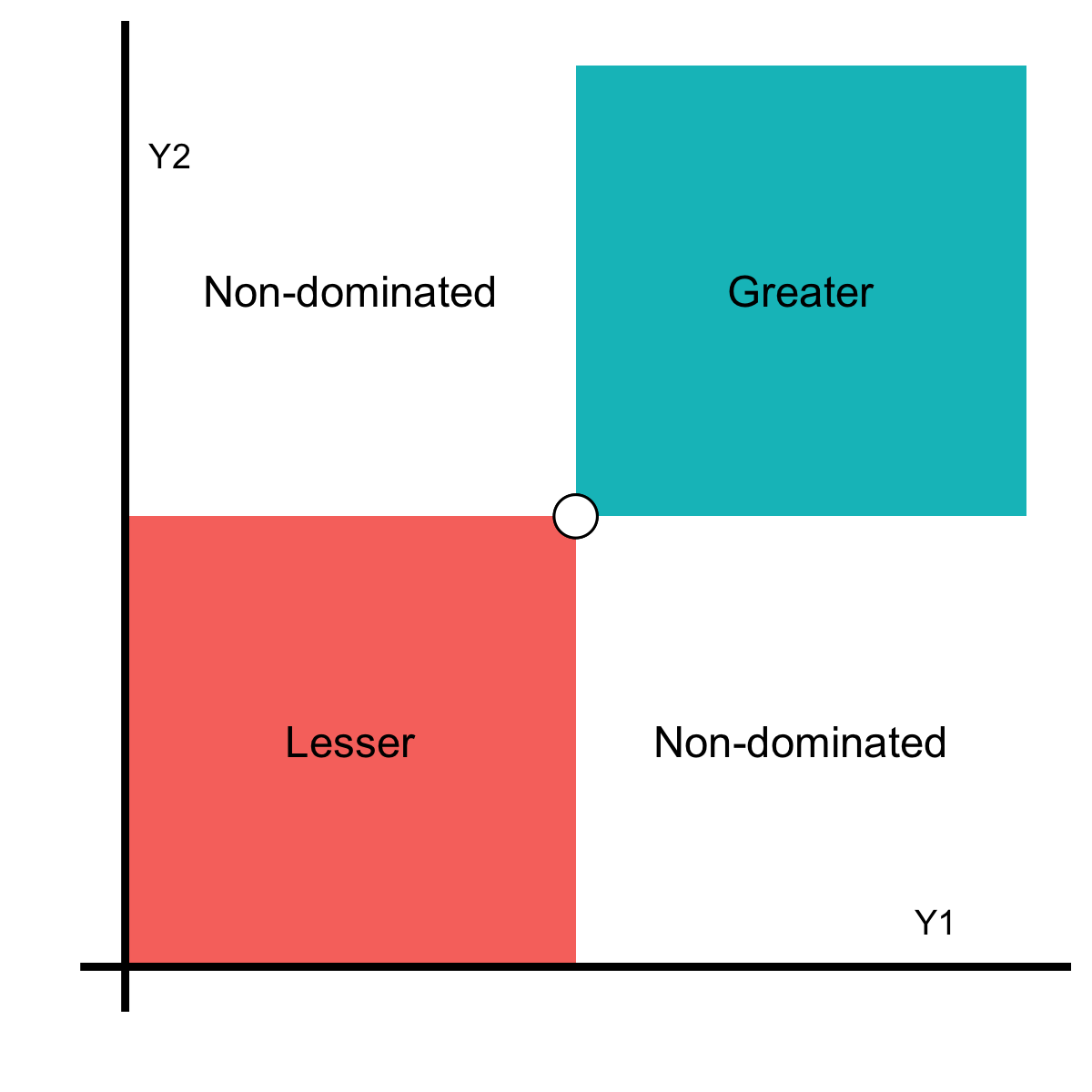}
  \end{minipage} %
  \begin{minipage}{0.45\textwidth}
    \includegraphics[width=0.95\textwidth]{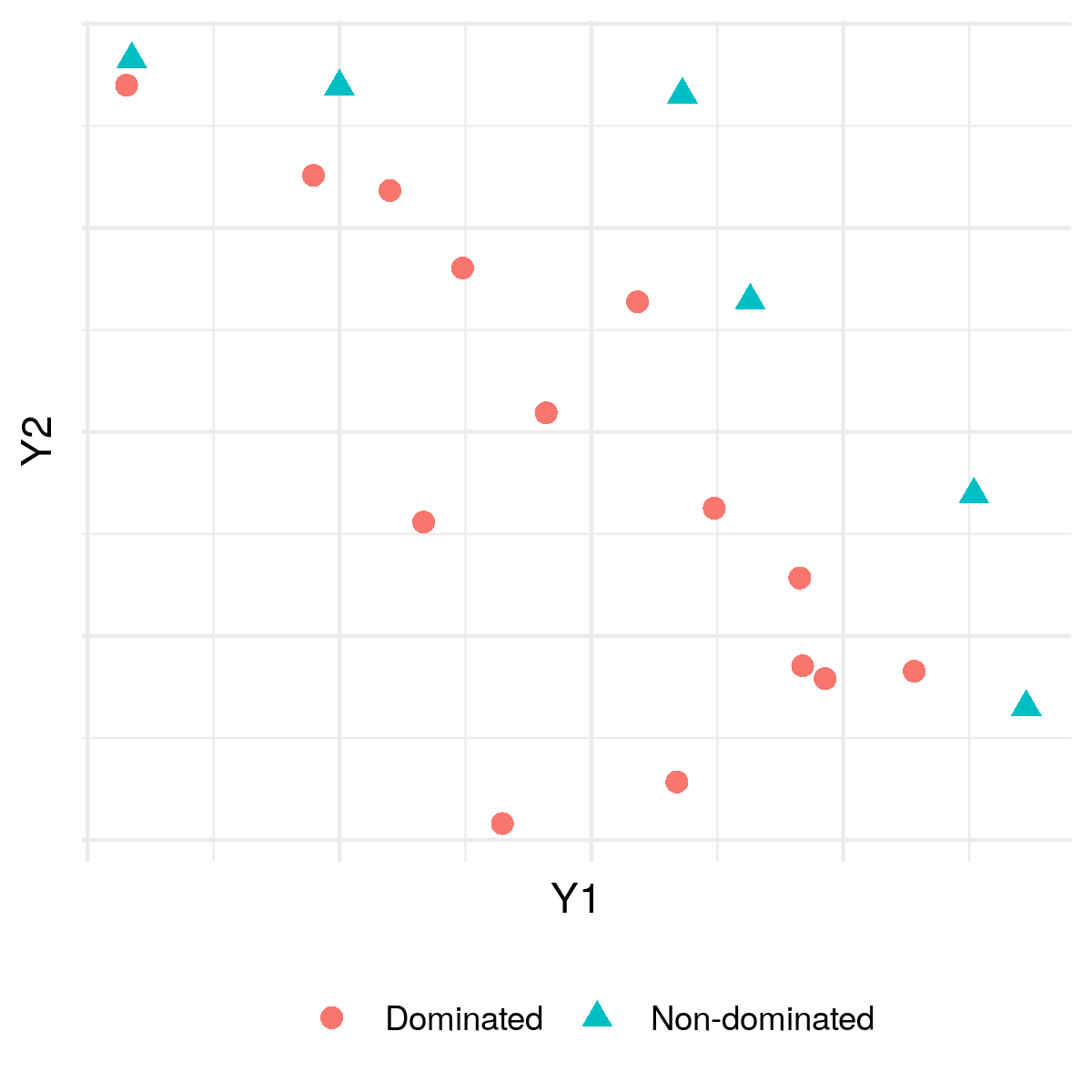}
  \end{minipage} %
  \caption{\emph{Multi-objective optimization requires ranking concepts beyond ``greater'' and ``lesser.''} Illustration of candidate ranking settings (left) and an example multi-objective frontier (right). In the single-objective setting, a relative ranking between candidates is always possible via the total ordering induced by the single objective. However, in the multi-objective setting, two candidates can be neither greater nor lesser than the other if they are alternately dominant along different objectives---two distinct candidates can be \emph{non-dominated}. Recognizing this issue leads to the notion of a \emph{non-dominated set} of mutually incomparable candidates---the Pareto frontier. (Figure inspired by Ref. \cite[Fig. 12.1]{kochenderfer2019algorithms}.)}
  \label{fig:ranking}
\end{figure}

In multi-objective optimization, the categories of ``greater'' and ``lesser'' are insufficient (Fig. \ref{fig:ranking}). Since two candidates can compete along multiple axes, it is possible for them to be mutually \emph{non-dominated}. Multiple objectives occur naturally in materials science problems and are often unavoidable, e.g. the strength-toughness trade-off which arises from fundamental, competing effects \cite{ritchie2011conflicts}. In lieu of more preference information, one must navigate the resulting multi-objective trade-off space. Having access to more candidates in this trade-off space enables more complete scientific understanding and more informed engineering decisions. In this section we review concepts from multi-objective optimization and introduce acquisition functions for the multi-objective setting.

\subsection{Dominance and strata}

In the single-objective case candidates $y, y'\in\R{}$ can be unambiguously ranked: A candidate $y$ is either lesser $y < y'$, greater $y > y'$, or equal $y = y'$ to another candidate $y'$. However, the introduction of multiple objectives $\vy, \vy' \in \R{D}$ introduces new complexities. Figure \ref{fig:ranking} illustrates the universe of possible comparisons; in addition to lesser, greater, and equal, in the multi-objective setting there exist \emph{non-dominated} points \cite[Chapter~12]{kochenderfer2019algorithms}. \emph{Dominance} (in the Pareto sense) is a pairwise relationship; the candidate $\vy$ is said to \emph{dominate} $\vy'$ if

\begin{equation} \begin{aligned} \label{eq:dominated}
                &y_d \geq y_d' \text{ for all } d \in \{1, \dots, D\}, \\
    \text{and } &y_d   >  y_d' \text{ for some } d.
\end{aligned} \end{equation}

\noindent We use the notation $\vy' \prec \vy$ to denote that $\vy$ dominates $\vy'$. Note that the definitions above pre-suppose that optimization is posed in terms of maximization of all objectives. This is not a limitation---if minimization is desired for a given axis $y_d$, then one must simply reverse the relevant inequalities in definition \ref{eq:dominated}. Furthermore, proximity to a desired value can be encoded as minimization of absolute distance from the target value, e.g. bandgap as close as possible to $1.2 eV$.

In the multi-objective setting, the possibility of non-dominance implies that a single ``best'' multi-objective output value $\vy$ may not exist. Instead, there may be a set of ``best values''. Given a set of candidates $\cA \subseteq \R{D}$, the set of non-dominated points is called the \emph{Pareto frontier}, defined by

\begin{equation}
  \cP(\cA) = \{\vy \in \cA \,|\, \forall \vy'\in\cA, \vy \not\prec \vy'\}.
\end{equation}

\noindent The Pareto frontier $\cP(\cA)$ represents the set of trade-offs one must navigate in choosing an optimization candidate. Further selection must be made through external considerations: possibly a prioritization of the objectives, or through harder-to-quantify concerns such as the corrosion resistance of a material \cite[Ch.~5]{ashby2011}.

While the Pareto frontier is an important set of candidates, points outside the frontier are not without utility, particularly if aforementioned external concerns exist. Candidates near the frontier can be useful as training data for a machine learning model, while measurement or model uncertainties may lead to the false classification of a point as dominated. To describe points near the Pareto frontier, we use the notion of \emph{strata}.

The \emph{strata} are defined via a recursive relationship \cite{roocks2016rPref}. Let $\cA$ be a set of candidates as above, and define the s-th stratum $\cS_s$ via

\begin{equation} \begin{aligned}
    \cS_1 &= \cP(\cA), \\
    \cS_s &= \cP(\cA - \cS_{s-1})\text{ for } s=2,\dots.
\end{aligned} \end{equation}

\begin{figure}
  \centering
  \begin{minipage}{0.45\textwidth}
    \includegraphics[width=0.95\textwidth]{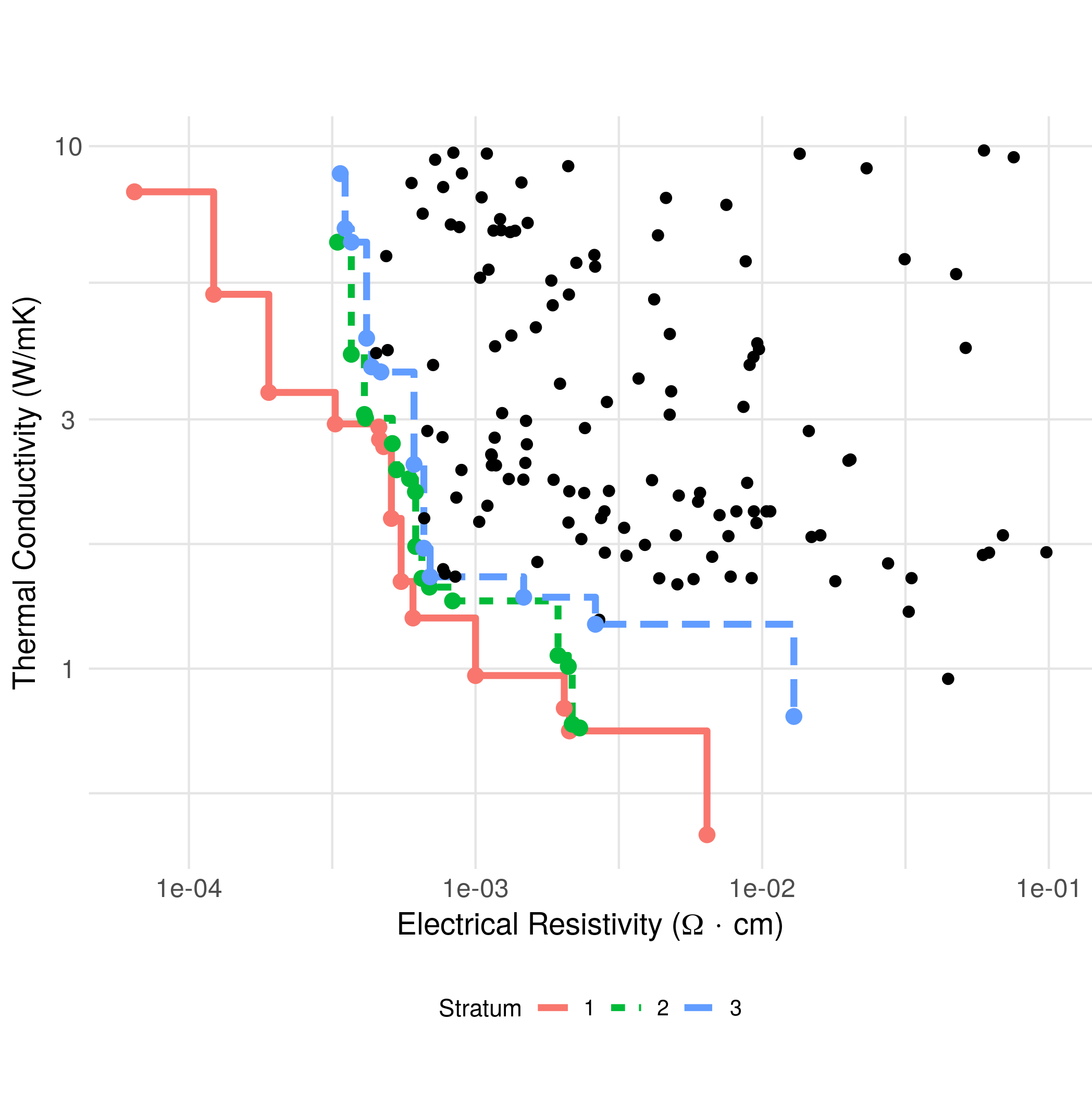}
  \end{minipage} %
  \begin{minipage}{0.45\textwidth}
    \includegraphics[width=0.95\textwidth]{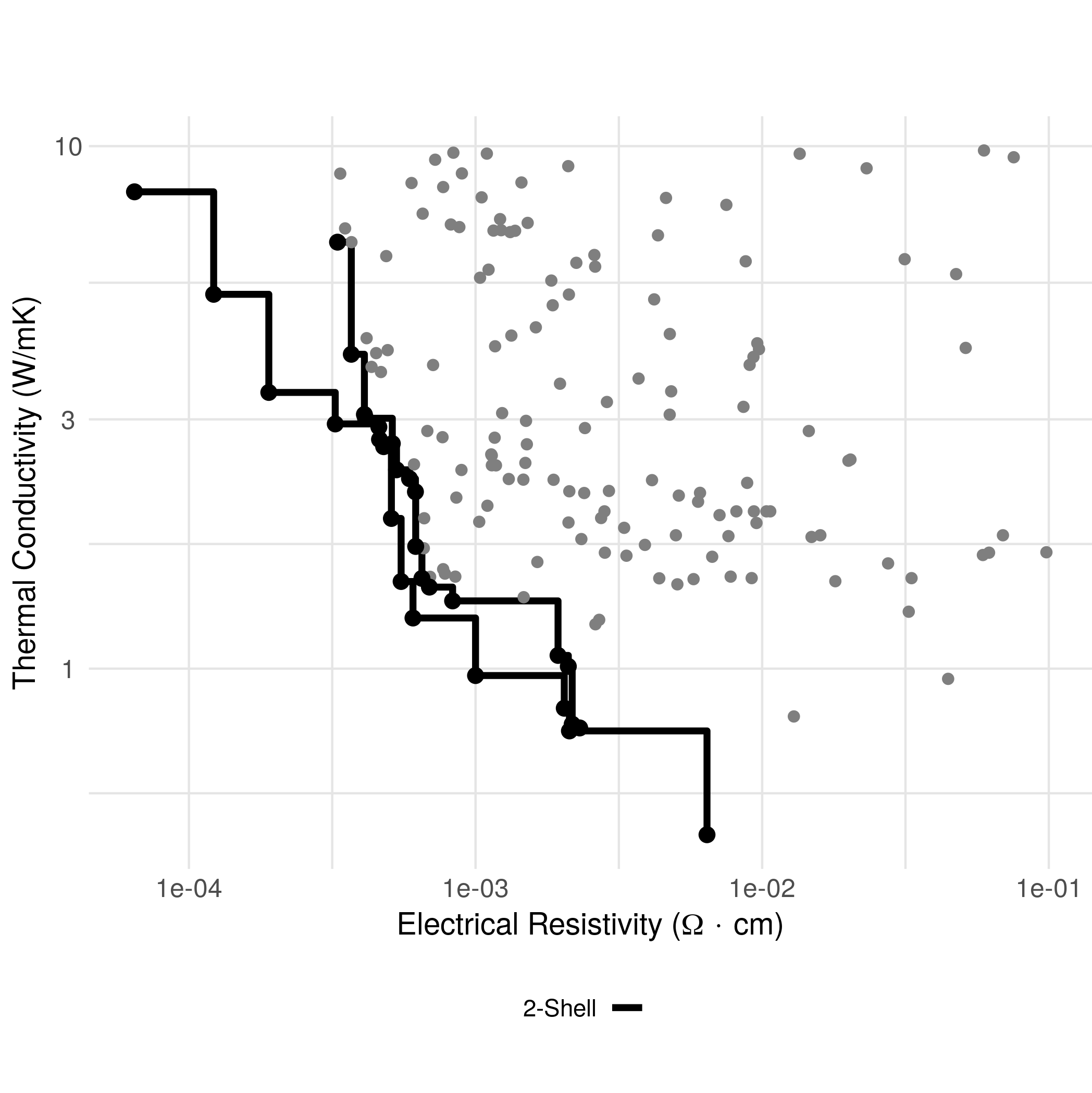}
  \end{minipage} %
  \caption{\emph{A recursive generalization of non-dominance (strata) enables a definition of scope tailored for optimal material candidates.} Example of strata (left) and a Pareto 2-shell (right) for the thermoelectric dataset, considering only $\rho, \kappa$ for preference ranking (both minimized). The 1-st stratum corresponds to the Pareto frontier, while higher strata are the Pareto frontier of the previous frontier's remainder (Def \eqref{eq:def-shell}). Considering all strata up to a number $s$ yields the Pareto $s$-shell. On the right, we visualize an example 2-shell, which consists of all the points in the first four strata. Below we use the concept of Pareto shells to define an error scope relevant to AL for materials discovery.}
  \label{fig:strata}
\end{figure}

\noindent Figure \ref{fig:strata} illustrates a few strata on a
thermoelectrics dataset (introduced in Subsection \ref{subsec:cases}). Note that by definition, we have $\cS_i \cap \cS_j =
\emptyset$ if $i\neq j$. This allows us to define a \emph{stratum number} for
each point $\vy \in \cA$ via

\begin{equation}
  s(\vy) = s\text{ if } \vy \in \cS_s.
\end{equation}

The candidates along the Pareto frontier then have $s(\vy) = 1$, while points with larger stratum number lie further from it. We will use the stratum number to rank the performance of AL with greater resolution than counting frontier points alone.

We also define the \emph{Pareto $s$-shell} via

\begin{equation} \label{eq:def-shell}
  \cP_s = \bigcup_{j=1}^s \cS_j.
\end{equation}

\noindent This definition allows one to select a ``band'' of points along and near the Pareto frontier. Below, we will use the Pareto shell $\cP_s$ as a targeted scope for computing model accuracy. We use the nomenclature $s$-shell to denote $\cP_s$: Figure \ref{fig:strata} illustrates a Pareto 2-Shell for the thermoelectric dataset.

\subsection{Measuring performance}

\paragraph{Candidate Improvement Performance}

Prior work assessing AL in materials informatics has focused primarily on the number of Pareto frontier points acquired during AL \cite{ling2017high,solomou2018multi}. While relevant, this ``all or nothing'' measurement of success provides no accounting for candidates near the Pareto frontier, which may still be of scientific interest. This is particularly troublesome when studying datasets that have very few Pareto frontier points, such as the ``sparse'' frontier we will consider below. To provide a measure with more granularity, we consider the \emph{mean stratum number} at each iteration of AL $\tilde{\cY}_{k}$, using the ground truth strata from the full dataset.

\paragraph{Model Accuracy Performance}

In addition to the candidate improvement performance of AL, we also consider trajectories of model performance. We use the notion of \emph{non-dimensional error} (NDE) across each of the output quantities; given a set of true $\{\vy_i\}_{i=1}^n = \cY$ and estimated $\{\hat{\vy}_i\}_{i=1}^n = \hat{\cY}$ response values, we compute the NDE for output $d$ via

\begin{equation} \label{eq:nde-def}
  \text{NDE}_d = \sqrt{ %
    \frac{ %
      \sum_{i=1}^n (y_{i,d} - \hat{y}_{i,d})^2
    }{ %
      \sum_{i=1}^n (y_{i,d} - \overline{y}_{d})^2
    } %
  },
\end{equation}

\noindent where $\overline{y}_d = \frac{1}{n} \sum_{i=1}^n y_{i,d}$ is the sample mean. If we have $D$ output quantities, then there are $D$ NDE values to compute. Note that the NDE is closely related to the \emph{coefficient of determination} $R^2$ via the expression $\text{NDE} = \sqrt{1 - R^2}$ \cite{weisberg2005applied}. Since the NDE is dimensionless, we may safely average NDE values across the $D$ outputs to compute a \emph{mean non-dimensional error} (MNDE), given by

\begin{equation} \label{def:mnde}
  \text{MNDE} = \frac{1}{D} \sum_{d=1}^D \text{NDE}_d.
\end{equation}

Given the retrospective nature of our test cases, the global error is naturally evaluated by \eqref{eq:nde-def} across the entire dataset. However, this \emph{global scope} includes regions of output space that are not emphasized in our frontier-seeking context. To compute error metrics with \emph{targeted scope}, we use the notion of a Pareto shell to compute error on the relevant portions of the output domain (Fig. \ref{fig:strata}). This notion of scope is closely related to the concept of a \emph{domain of applicability}, with the contrast that the scope is defined by the analyst, whereas the aforementioned domain is identified from a trained machine learning model \cite{sutton2019identifying}.

\subsection{Acquisition Functions} \label{subsec:acq-fcn}

To address the multivariate nature of the material properties $\vy \in \R{D}$, we introduce decision criteria that summarize available information about material candidates and provide a ranking. To that end, we define an \emph{acquisition function} $f_a(\vx; \cM)$ as any function which is used to select an optimal candidate via

\begin{equation} \label{eq:acq-general}
  \vx^* = \argmax_{\vx \in \tilde{\cX}^{C}} f_a(\vx; \cM)
\end{equation}

\noindent where $\tilde{\cX}^{C}$ is the complement of the training set, and $\cM$ is a trained machine learning model which returns both a prediction $\hat{\vy}$ and a predictive distribution $\mY\sim\hat{\cD}$, both of which are available to the acquisition function $f_a$.

Many generalizations of single-objective acquisition functions are available from the literature, including the probability of improvement and expected improvement criteria \cite{keane2006statistical}, the max-min criterion \cite{svenson2016multiobjective}, and the expected hyper-volume improvement (EHVI) criterion \cite{solomou2018multi}. Comparing these and other criteria is outside the scope of this work; instead, we are focused on (i) how the fidelity with which the Pareto frontier is represented relates to AL performance, and (ii) how optimal candidate selection is (or is not) related to model accuracy. Below, we introduce the acquisition functions to be studied in this work, after a remark on dimensional homogeneity.

\paragraph{Importance of dimensional homogeneity}

Before introducing specific acquisition functions, we first make a remark on the importance of \emph{dimensional homogeneity}. In order to measure the performance of multi-objective AL, we may wish to measure the ``distance'' of candidates to the Pareto frontier. However, we will see here that a naive notion of distance is problematic, as it does not respect dimensional homogeneity---the ranking results are not independent of the analyst's choice of unit system.

To illustrate, let $\vy,\vy' \in \R{D}$, where the $y_d$ potentially have different physical units. The ordinary notion of distance is given by

\begin{equation} \label{eq:p-norm}
  \|\vy - \vy'\|_p = \left( \sum_{d=1}^D |y_d - y'_d|^p \right)^{1/p}.
\end{equation}

\noindent Note that for all $p \in (0, +\infty)$, this expression involves the addition of terms of potentially different physical units. This expression violates dimensional homogeneity, which introduces an artificial dependence on the chosen unit system. Thus the ranks computed by distances are not necessarily stable to a change of unit system---we provide an example of this pathology in the Supplementary Material. This illustrates that arbitrary choices can drastically affect decisions based on this sort of distance computation. To overcome this issue, we only consider acquisition functions that respect dimensional homogeneity.

\paragraph{Uncertainty sampling}

On the scale from exploitation to exploration, \emph{uncertainty sampling} leans heavily towards the latter; one chooses candidates based on where the model is most uncertain. In the scalar case, this is easily accomplished by choosing the candidate with the greatest predictive variance $\hat{\sigma}_i^2$ \cite{settles2009active}. This approach does not immediately generalize to the multi-objective case, as the component variances $\hat{\sigma}_{i,d}^2 = \hat{\m\Sigma}_{i,dd}$ do not necessarily have the same units. To respect dimensional homogeneity, we generalize uncertainty sampling by considering a sum of dimensionless quantities.

\emph{Sum of Coefficients of Variation} (SCV): The SCV is defined via

\begin{equation} \label{eq:def-scv}
  f_{\text{SCV}}(\vx_i) = \sum_{d=1}^D \text{COV}_{i,d},
\end{equation}

\noindent where the $\text{COV}_{i,d} = \sigma_{i,d} / |\mu_{i,d}|$ are coefficients of variation, and $\hat{\mu}_{i,d}, \hat{\sigma}_{i,d}$ are the $d = 1, \dots, D$ components of the mean and standard deviation of the predictive distribution $\hat{\cD}_i$. Note that this definition is problematic if any of the $\mu_i$ are exactly zero. However, the coefficients of variation $\text{COV}_{i,d}$ are dimensionless quantities, with a normalizing scale $\hat{\mu}_{i,d}$ set not by the user, but rather by the available data.

\paragraph{Frontier modeling strategies}

Here we introduce a family of acquisition functions that seek improvement over an existing Pareto frontier, in order of increasing fidelity with which they model the Pareto frontier. Each of these strategies is a form of probability statement; by construction, these quantities respect dimensional homogeneity.

\begin{figure}
  \centering
\begin{minipage}{0.32\textwidth}
  \includegraphics[width=0.95\textwidth]{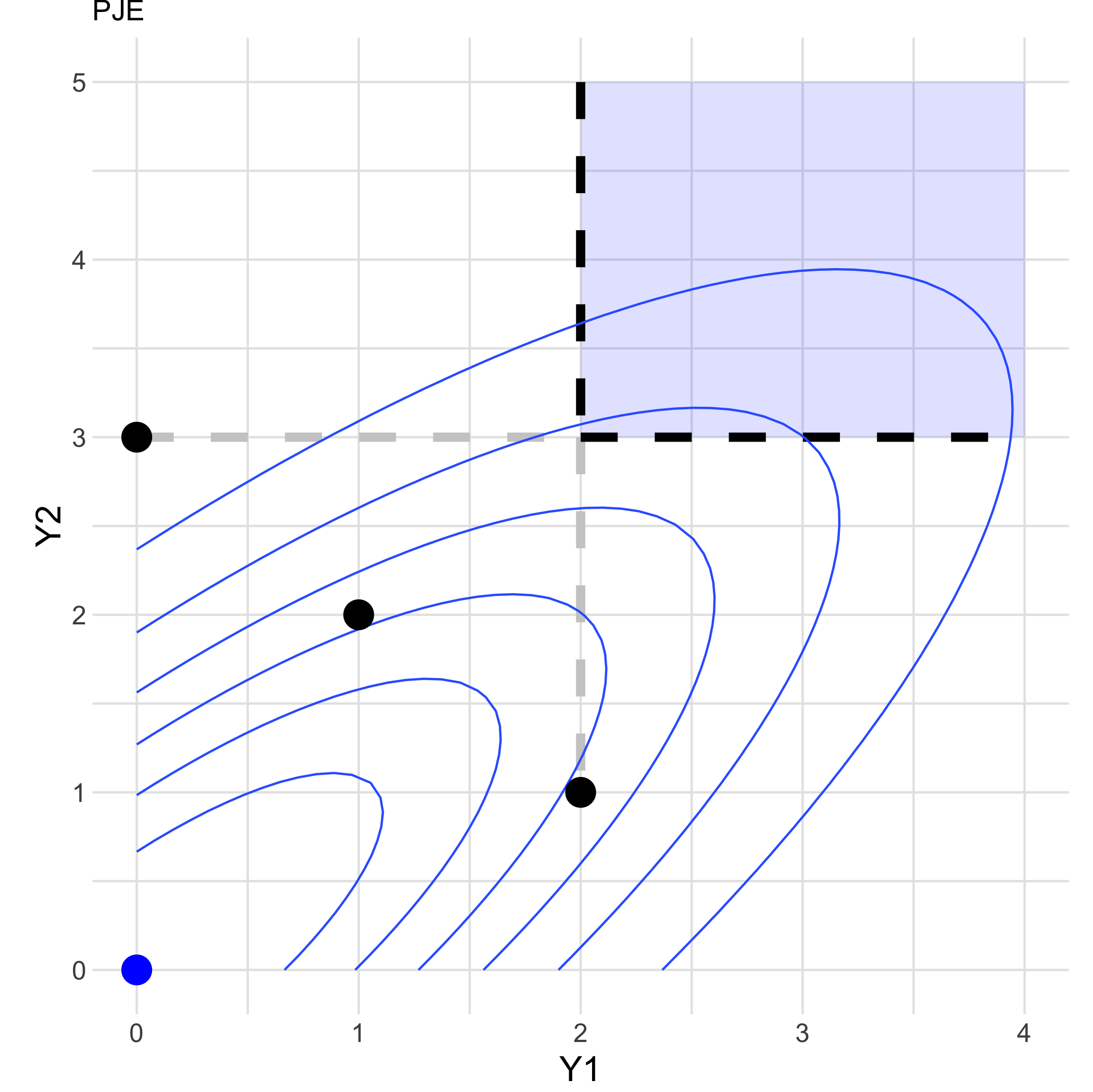}
\end{minipage} %
\begin{minipage}{0.32\textwidth}
  \includegraphics[width=0.95\textwidth]{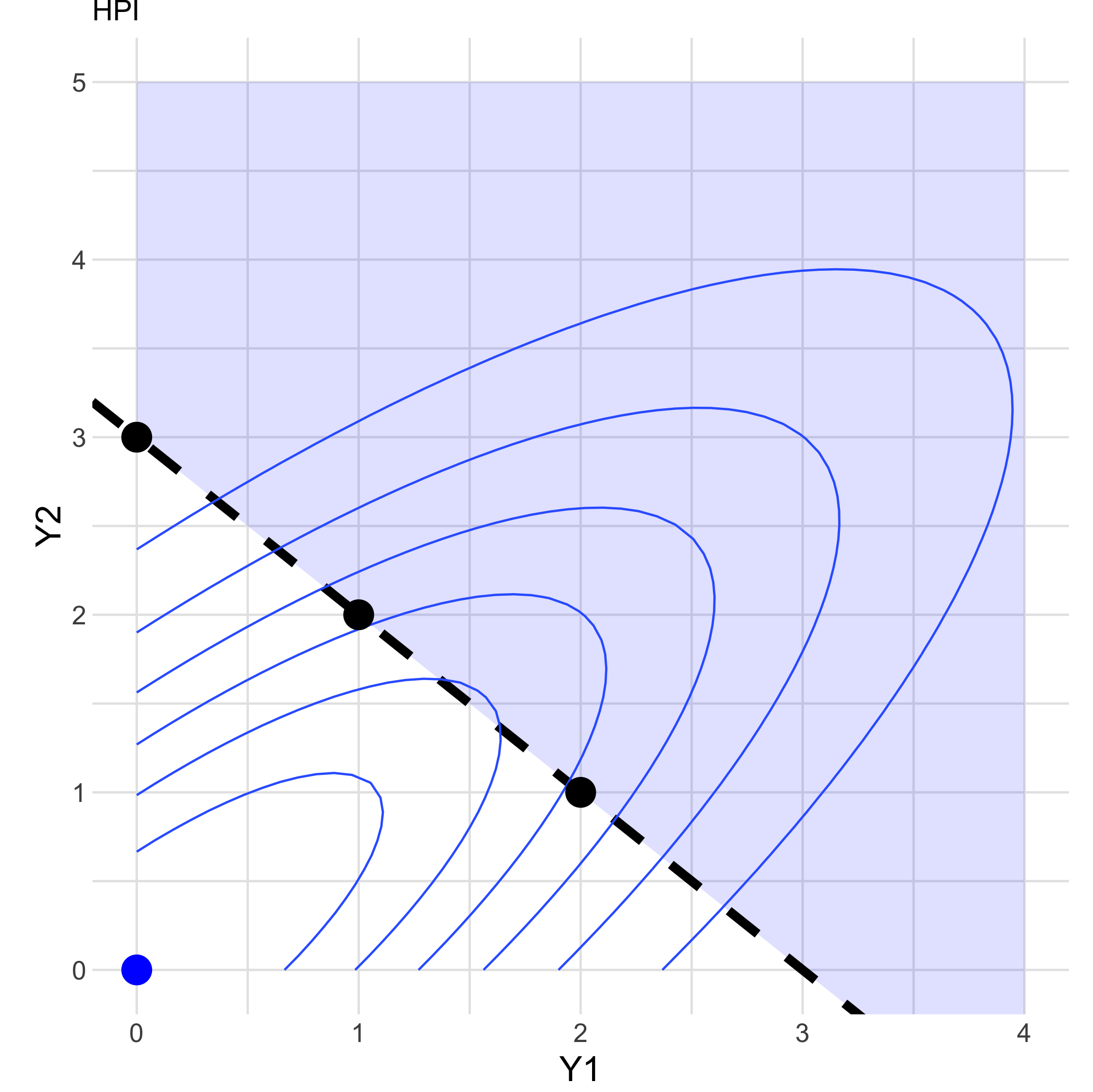}
\end{minipage} %
\begin{minipage}{0.32\textwidth}
  \includegraphics[width=0.95\textwidth]{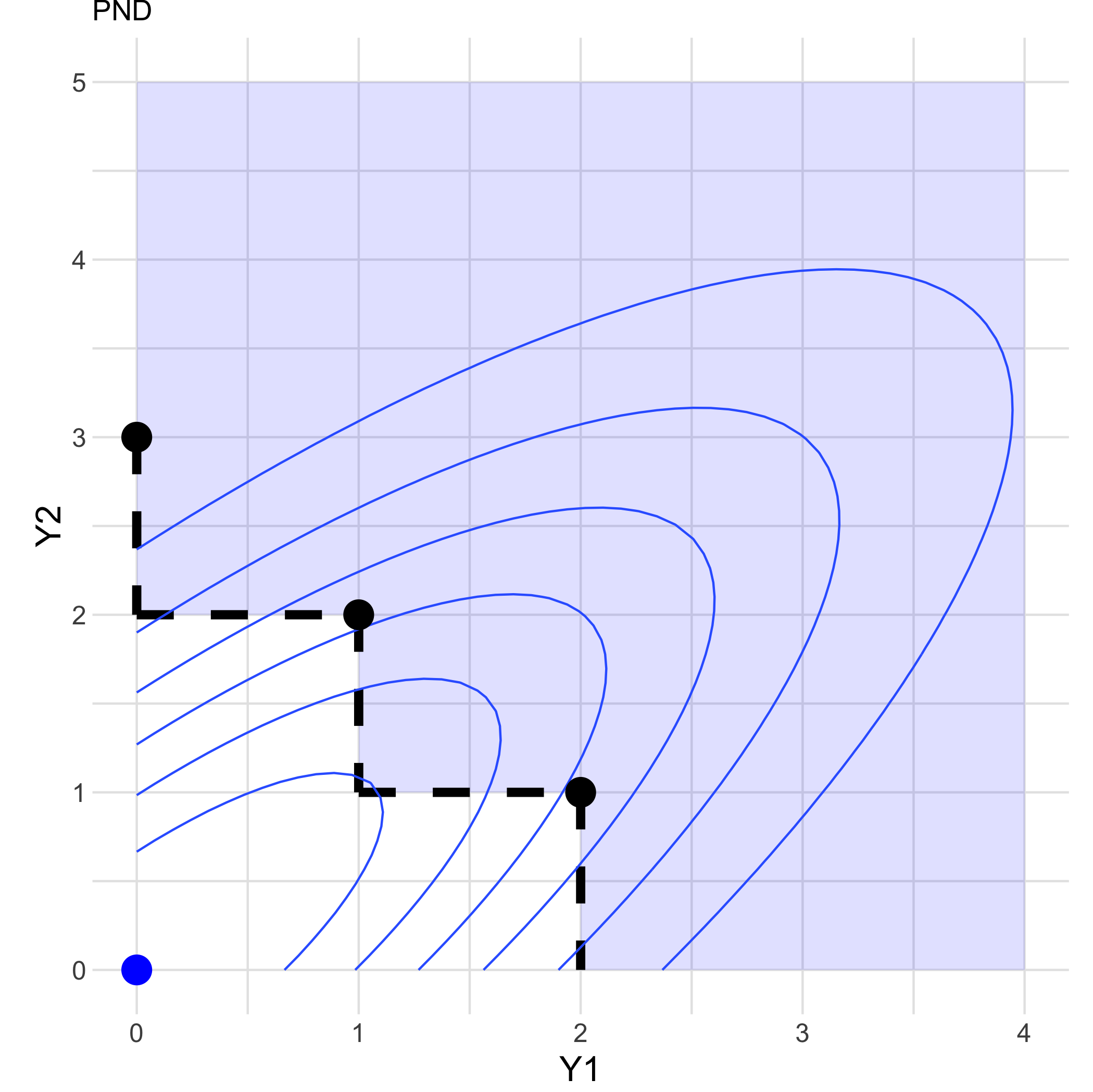}
\end{minipage} %
  \caption{\emph{Schematics illustrating acquisition functions} probability of joint exceedence (PJE, left), hyperplane probability of improvement (HPI, center), and probability non-dominated (PND, right) for the same candidate (blue, at origin) and frontier points (black). Both outputs $Y1,Y2$ are to be maximized. The blue curves depict equi-likelihood contours for a single candidate's predictive density $\hat{\cD}$. The shaded region depicts the region that is integrated for the respective acquisition function. Note that PJE largely ignores the frontier geometry, HPI crudely models the Pareto frontier, and PND accurately considers the Pareto frontier.}
  \label{fig:def-acq}
\end{figure}

\emph{Probability of Joint Exceedance} (PJE): The PJE is defined via

\begin{equation} \label{eq:def-pje}
  f_{\text{PJE}}(\vx_i) = \P_{\cD_i}[Y_{i,d} > \max_{\vy \in \tilde{\cY}} y_d],
\end{equation}

\noindent where $\mY_i \sim \hat{\cD}_i$ is the random variable which follows the predictive distribution $\hat{\cD}_i$ for candidate $i$. In words, definition \eqref{eq:def-pje} is the probability that candidate $\mY_i$ will exceed the performance observed in our existing training data $\tilde{\cY}$ along every axis of comparison. This is a very ``aggressive'' acquisition function that ignores much of the structure of the Pareto frontier. Figure \ref{fig:def-acq} illustrates the PJE, alongside the other frontier modeling criteria.

\emph{Hyperplane Probability of Improvement} (HPI): The HPI is defined in terms of a hyperplane fit of the Pareto frontier in the available training data. Modeling the Pareto frontier as a hyperplane requires fitting a normal direction $\hat{\vw} \in \R{D}$ and an offset $\hat{b}$. Once these are fit, the HPI is defined in terms of the appropriate Z-score via

\begin{equation}
  f_{\text{HPI}}(\vx_i) = \frac{ %
    \hat{\vw}^{\top} \hat{\v\mu}_i - \hat{b} %
  }{ %
    \sqrt{ \hat{\vw}^{\top} \hat{\m\Sigma}_i \hat{\vw} }
  }.
\end{equation}

\noindent In words, the HPI is the probability that a candidate $\mY_i$ will present improvement over the hyperplane fit of the existing Pareto frontier.

There are many ways to fit a hyperplane to a set of data. One option is to arbitrarily select one output $Y_{i}$ and perform linear regression using the remaining outputs as regression variables. We recommend against this approach, as it suffers from the \emph{regression fallacy} \cite[Ch.~9.1]{owen2013regression}. In practice, we perform a principal component analysis of the available Pareto frontier data, and use the least-variance direction to define the hyperplane direction $\hat{\vw}$. Together with the mean of the Pareto frontier data $\overline{\mY}$, we then define the offset via $\hat{b} = \hat{\vw}^{\top} \overline{\mY}$.

Figure \ref{fig:def-acq} illustrates the HPI: Note that this definition assumes a hyperplane structure, which will not be appropriate for all Pareto frontiers. Further, the HPI ``fills in'' the staircase structure posed by the true Pareto frontier---the final frontier modeling acquisition function captures this structure.

\emph{Probability Non-Dominated} (PND): The PND is defined via

\begin{equation} \label{eq:def-pnd}
  f_{\text{PND}}(\vx_i) = \P_{\cD_i}[Y_{i} \not\prec \vy,\, \forall \vy \in \tilde{\cY}],
\end{equation}

\noindent with dominance $\prec$ defined in \eqref{eq:dominated}. In words, the PND computes the probability that a given candidate $\mY_i$ will be non-dominated with respect to the available data. This criterion is studied in approximate form in Reference \citenum{keane2006statistical}. The PND fully considers the known Pareto frontier, introducing no modeling assumptions. Figure \ref{fig:def-acq} illustrates this acquisition function against the aforementioned definitions.

Note that while PND captures the true geometry of the Pareto frontier, since the frontier is allowed to have quite general structure, no simple analytic expression as for PJE or HPI exists. Instead, one may approximate the PND via ordinary Monte Carlo, drawing samples from the predictive distribution $\cD_i$ and counting the proportion that are non-dominated. This leads to a greater computational expense to evaluate the PND, as compared with PJE or HPI. The experiments below will demonstrate that this added expense can be valuable if higher AL performance is desired.

\section{Results} \label{sec:results}

\subsection{Test cases} \label{subsec:cases}

\begin{figure}
  \centering
  \includegraphics[width=0.75\textwidth]{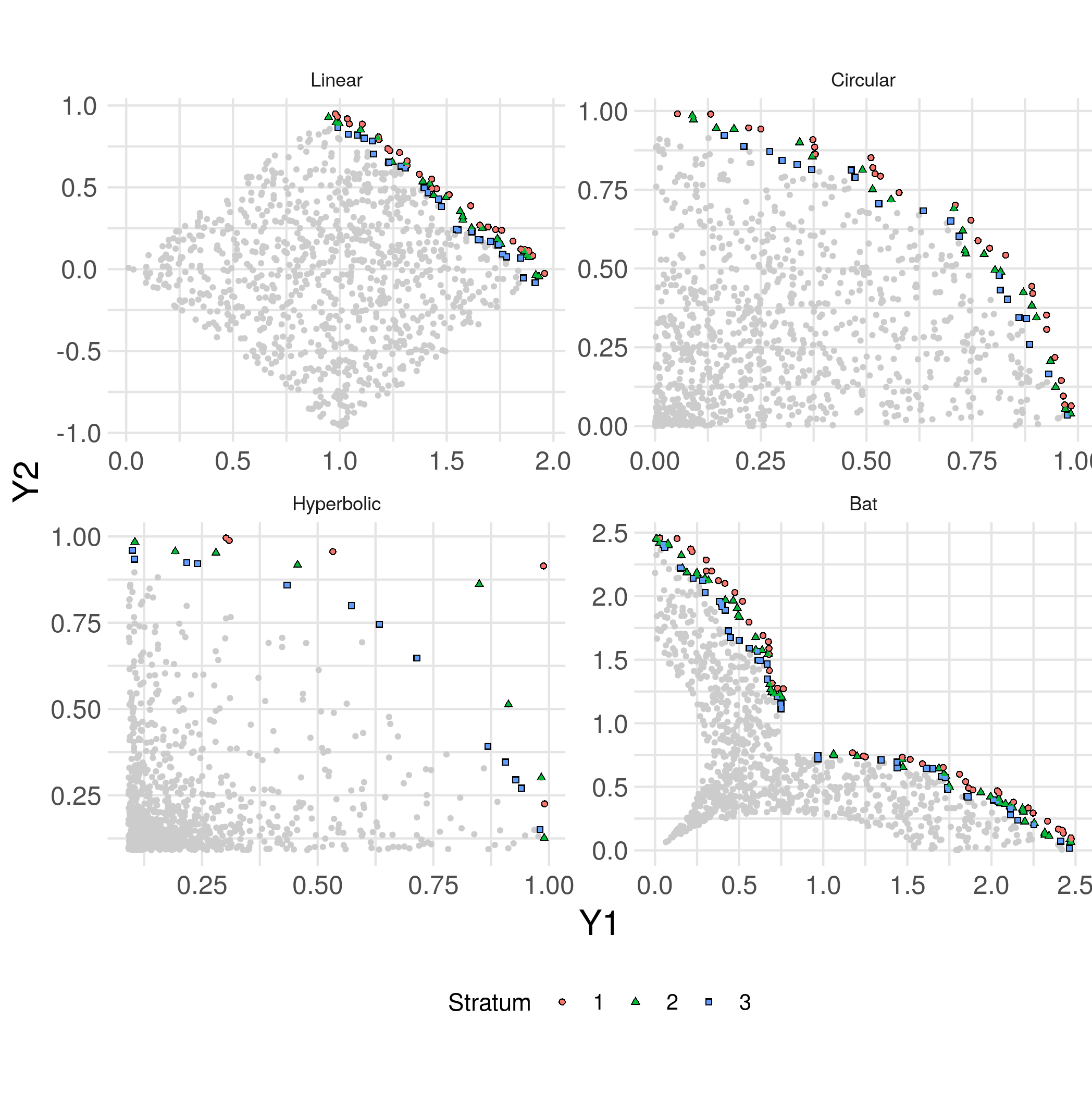}
  \caption{\emph{Synthetic test cases} in terms of their output (property) spaces. Each output space is two-dimensional, with both outputs to be maximized. Note that each test case has a different Pareto frontier geometry, including convex (\texttt{Circular}), concave (\texttt{Bat}), and sparse (\texttt{Hyperbolic}) examples.}
  \label{fig:synthetic-frontier-lineup}
\end{figure}

To compare the acquisition functions introduced above, we simulate AL on a collection of synthetic and experimental databases. The synthetic cases are constructed to present different Pareto frontier geometries, including a linear frontier, as well as examples of convex and concave frontiers. We also construct a ``sparse'' frontier containing relatively few low-stratum points. Figure \ref{fig:synthetic-frontier-lineup} presents these test-cases' two-dimensional output spaces. The functional forms for these models are given in Appendix \ref{apx:synthetic}.

\begin{figure}
  \centering
  \includegraphics[width=0.75\textwidth]{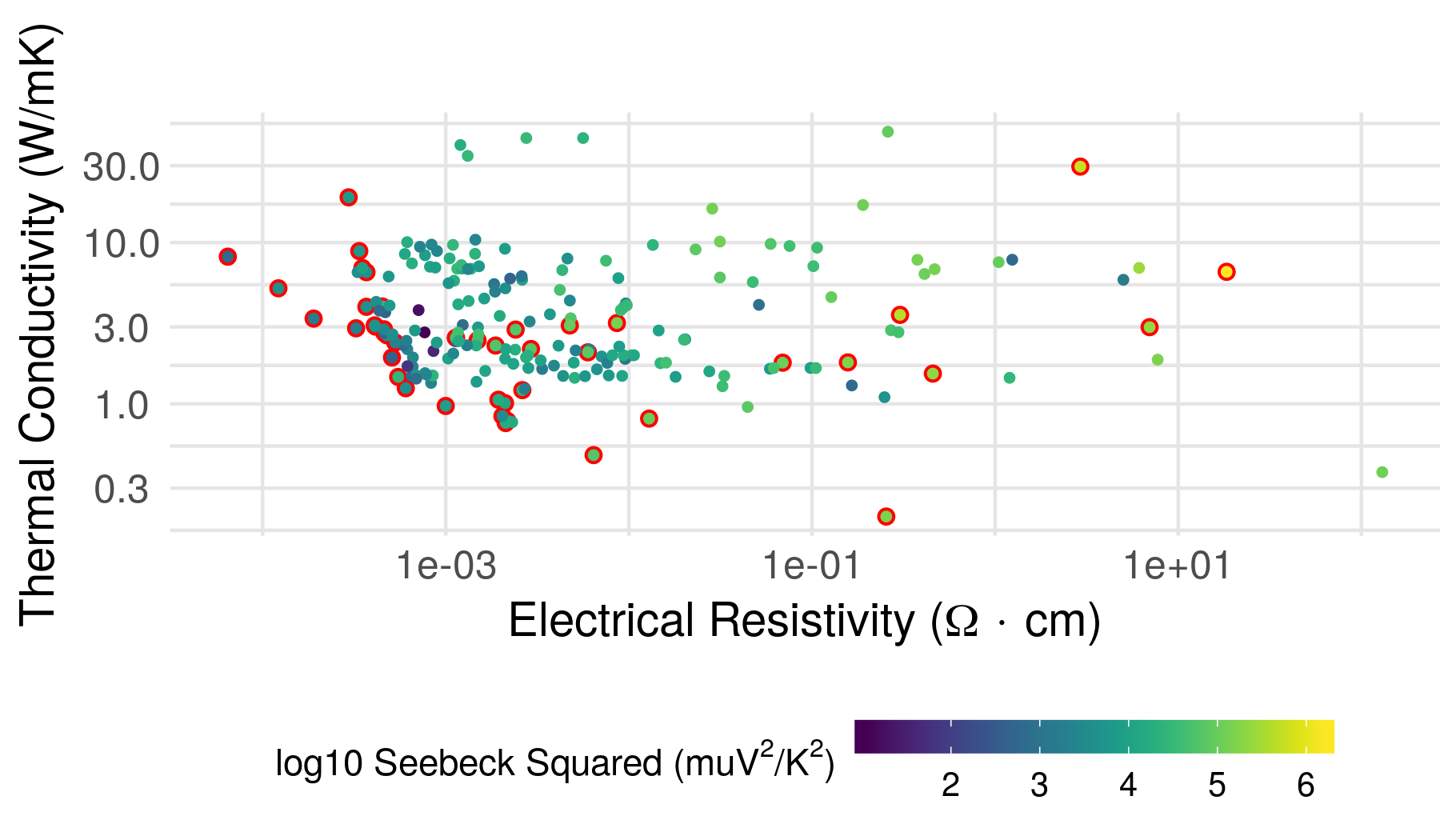}
  \caption{\emph{Thermoelectric dataset's output (property) space}. Outputs Thermal Resistivity and Electrical Conductivity are to be minimized, while the squared-Seebeck coefficient is to be maximized. Pareto frontier points are highlighted in red.}
  \label{fig:thermoelectric-frontier}
\end{figure}

We also consider a published dataset of thermoelectric materials \cite{gaultois2013data}, depicted in Figure \ref{fig:thermoelectric-frontier}. The performance (property) space consists of thermal conductivity $\kappa$, electrical resistivity $\rho$, and the Seebeck coefficient $S$. The inputs are computed using the Magpie featurization library: We use the \texttt{matminer} package to compute features (descriptors) including stoichiometry, valence orbital, and ion properties as inputs \cite{ward2016general,Ward2018MatminerAO}. Full code to load this dataset and featurization is available in the Supplementary Material.

\subsection{Retrospective active learning experiments}

The primary evidence of this work is based on a battery of AL simulations, which support the comparison of different acquisition functions against a common setup. For all experiments, we use the random forest model implemented in the \texttt{lolo} package \cite{hutchinson2016lolo}.

Chance phenomena such as initial data selection can affect AL results, implying that single runs of AL are insufficient for measuring performance. We consider an ensemble of runs against a randomized selection of initial data, in order to provide a robust estimate of relative performance.

To perform a single run of AL, we carry out the following steps:

\begin{enumerate}
\item Choose an initial random subset $\tilde{\cX}_{0} \subseteq \cX$ of size
  $|\tilde{\cX}_{0}| = C$ for the training data, and reveal the paired labels
  $\tilde{\cY}_{0}$. Set the iteration counter to $k=0$.
\item Fit a random forest model to the available paired data $\tilde{\cX}_{k},
  \tilde{\cY}_{k}$. This returns predictions $\hat{\vy}_i$ and predictive densities
  $\hat{\cD}_i$. \label{step:repeat}
\item Rank the candidates $\vx_i \in (\cX - \tilde{\cX}_{k})$ remaining in the dataset according to the chosen acquisition function $f_a(\vx_i)$. Select the top-performing candidate $\vx^*$, and add it to the database $\tilde{\cX}_{k+1} = \tilde{\cX}_{k} + \{\vx^*\}$. Reveal the label for $\vx^*$.
\item Repeat from Step \ref{step:repeat} for $k = 1, \dots, K-1$ total iterations.
\end{enumerate}

To perform an ensemble of AL runs, we select different random subsets $\tilde{\cX}_{0,1}, \dots, \tilde{\cX}_{0,R}$ for $R$ repetitions, and aggregate results across these runs. We vary the initial candidate pool size $C$, the total number of iterations $K$, the considered test case, and the chosen acquisition function. We performed $150$ active learning runs for each choice of acquisition function and test case: The Supplementary Material provides numerical details on $C,K,R$.

\subsection{Candidate improvement} \label{subsec:candidate}

\begin{figure}
  \centering
  \begin{minipage}{0.45\textwidth}
    \includegraphics[width=0.95\textwidth]{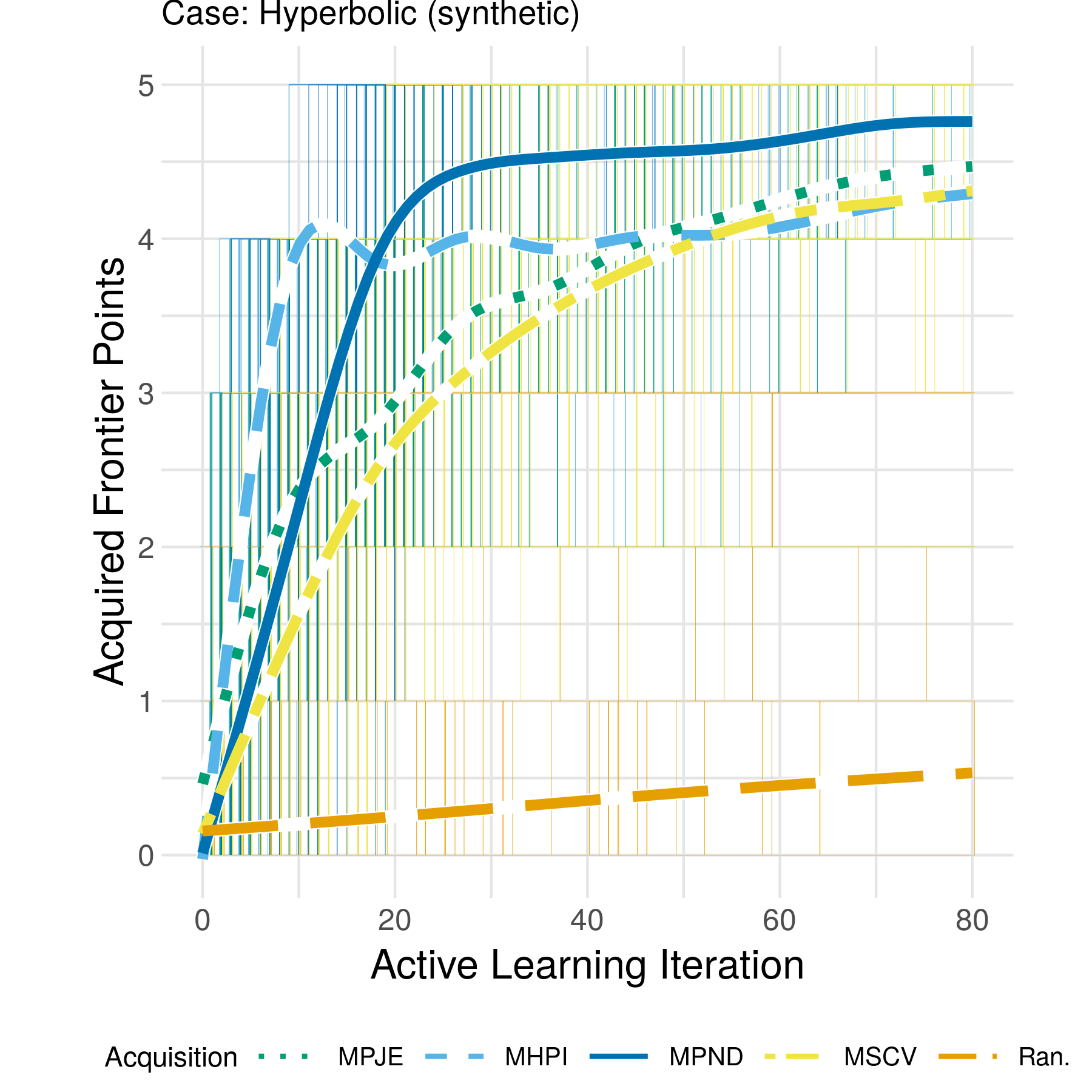}
  \end{minipage} %
  \begin{minipage}{0.45\textwidth}
    \includegraphics[width=0.95\textwidth]{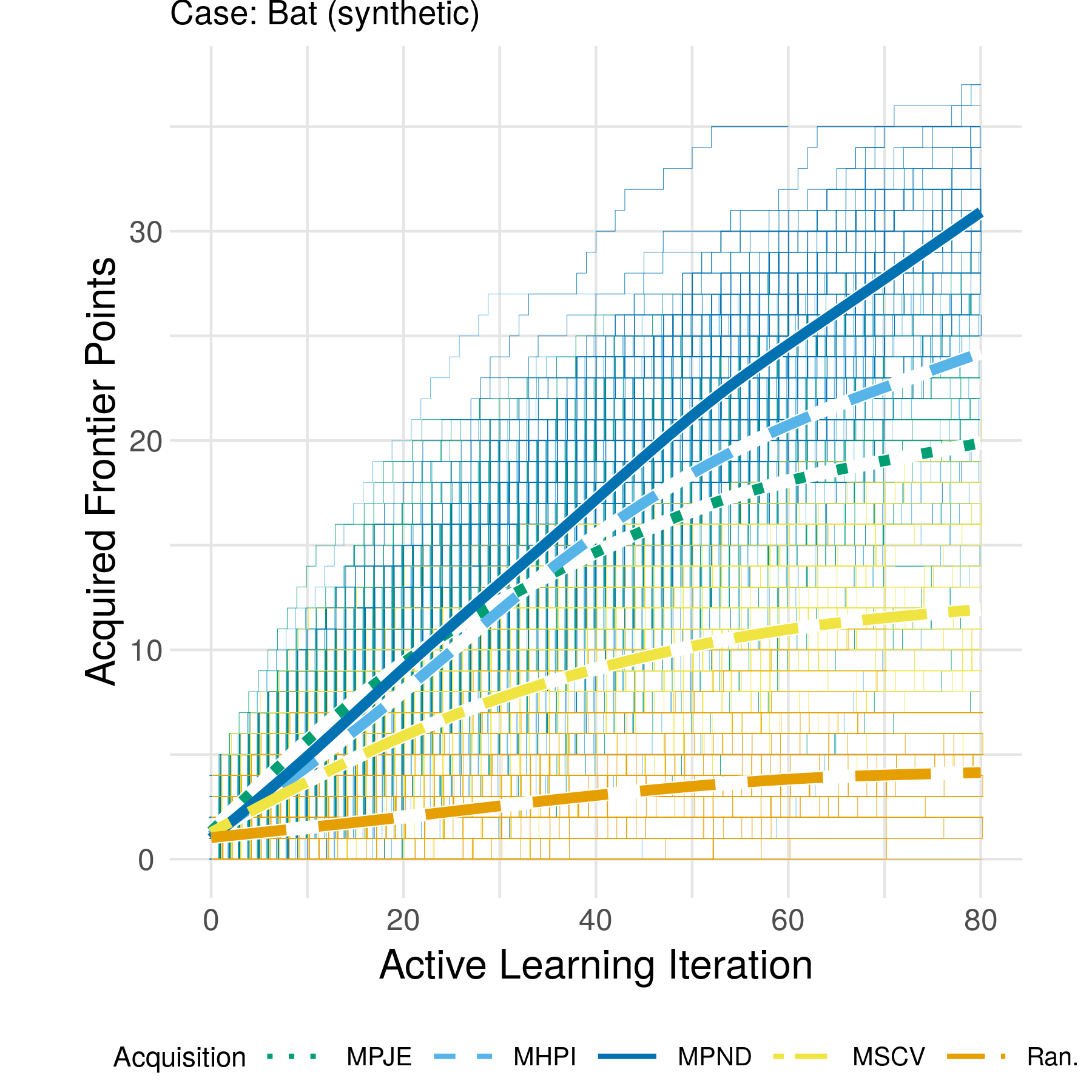}
  \end{minipage} %
  \caption{\emph{Greater acquisition function fidelity leads to more non-dominated candidates.} Total frontier points acquired for the \texttt{Hyperbolic} (left) and \texttt{Bat} (right) synthetic frontiers. Uncertainty sampling (MSCV) has unreliable performance, exhibiting mediocre performance on the \texttt{Bat} frontier, but performing well on the \texttt{Hyperbolic} test case. In terms of long-run performance, the criteria MPJE, MHPI, and MPND tend to rank in the same order as the fidelity with which they represent the Pareto frontier.}
  \label{fig:syn-total-nndp}
\end{figure}

\begin{figure}
  \centering
  \begin{minipage}{0.45\textwidth}
    \includegraphics[width=0.95\textwidth]{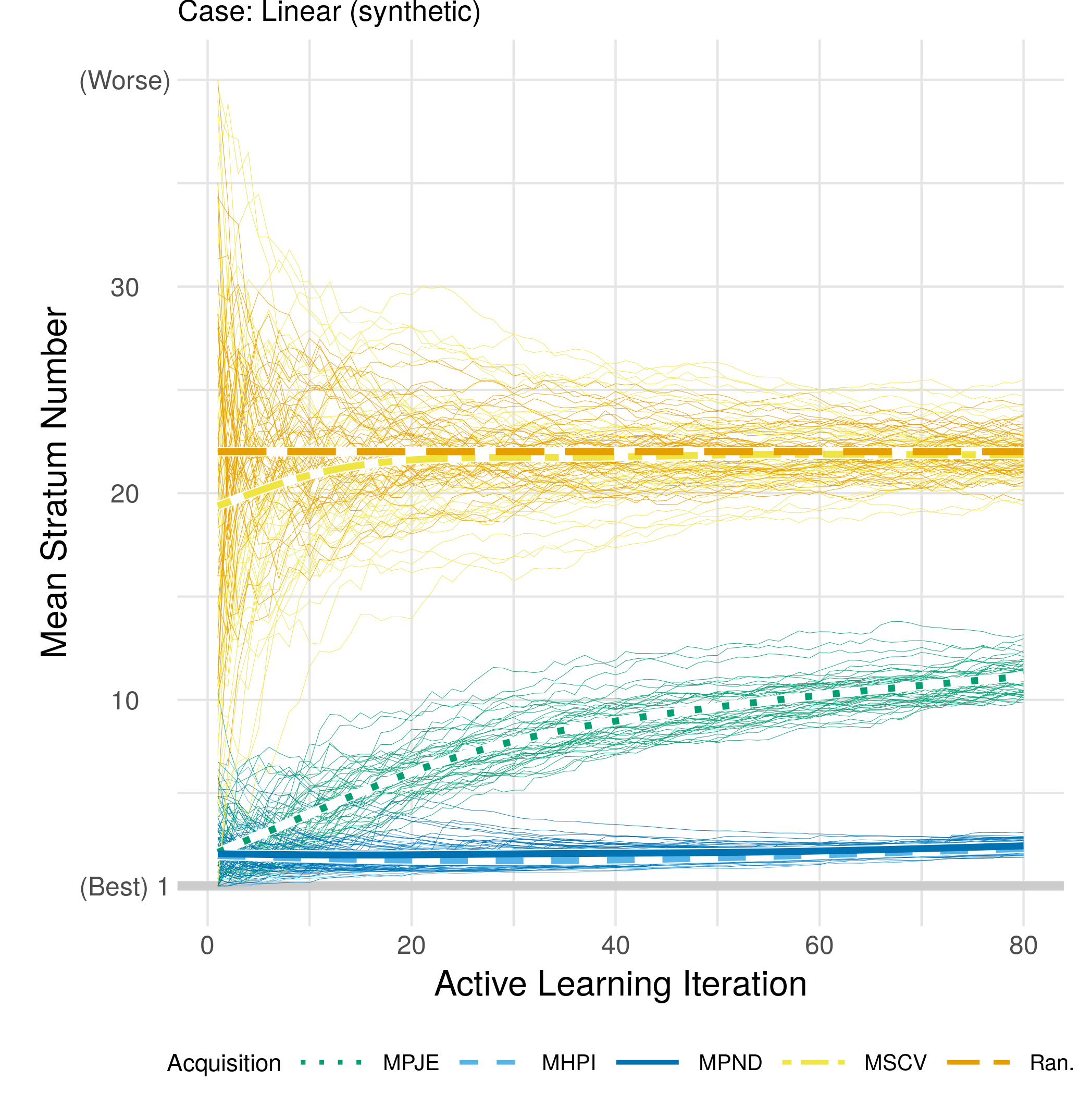}
  \end{minipage} %
  \begin{minipage}{0.45\textwidth}
    \includegraphics[width=0.95\textwidth]{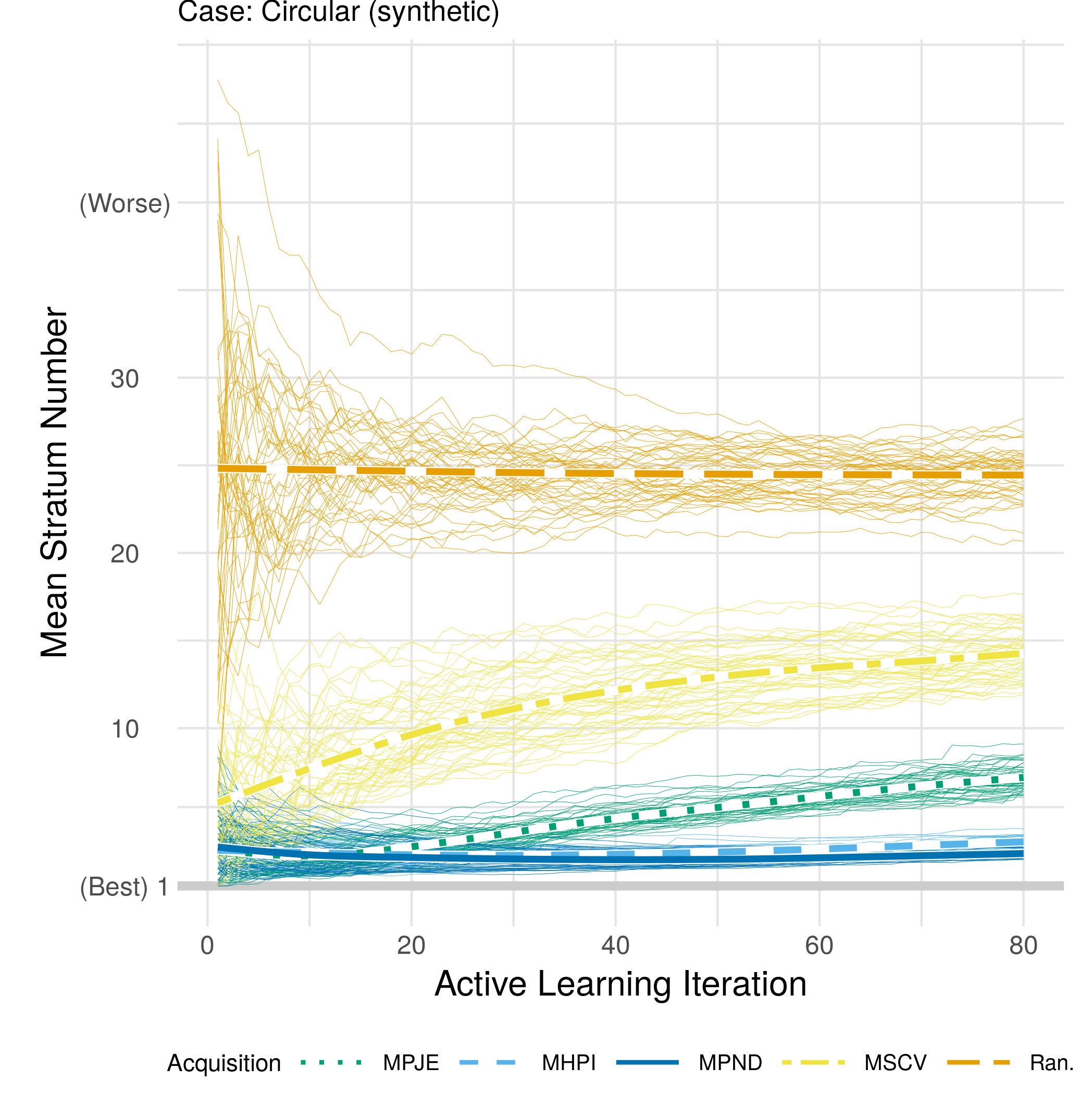}
  \end{minipage}
  \begin{minipage}{0.45\textwidth}
    \includegraphics[width=0.95\textwidth]{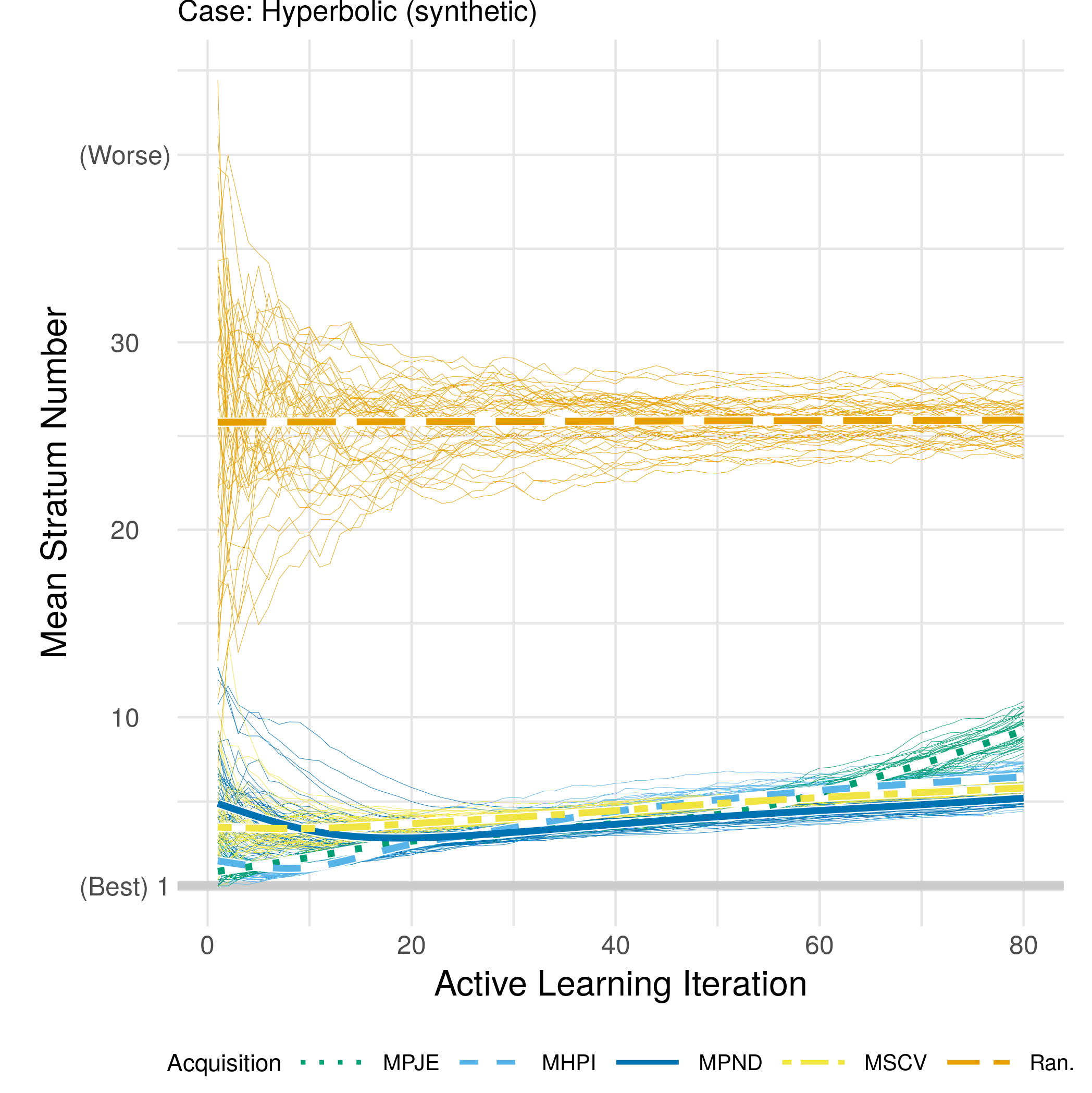}
  \end{minipage} %
  \begin{minipage}{0.45\textwidth}
    \includegraphics[width=0.95\textwidth]{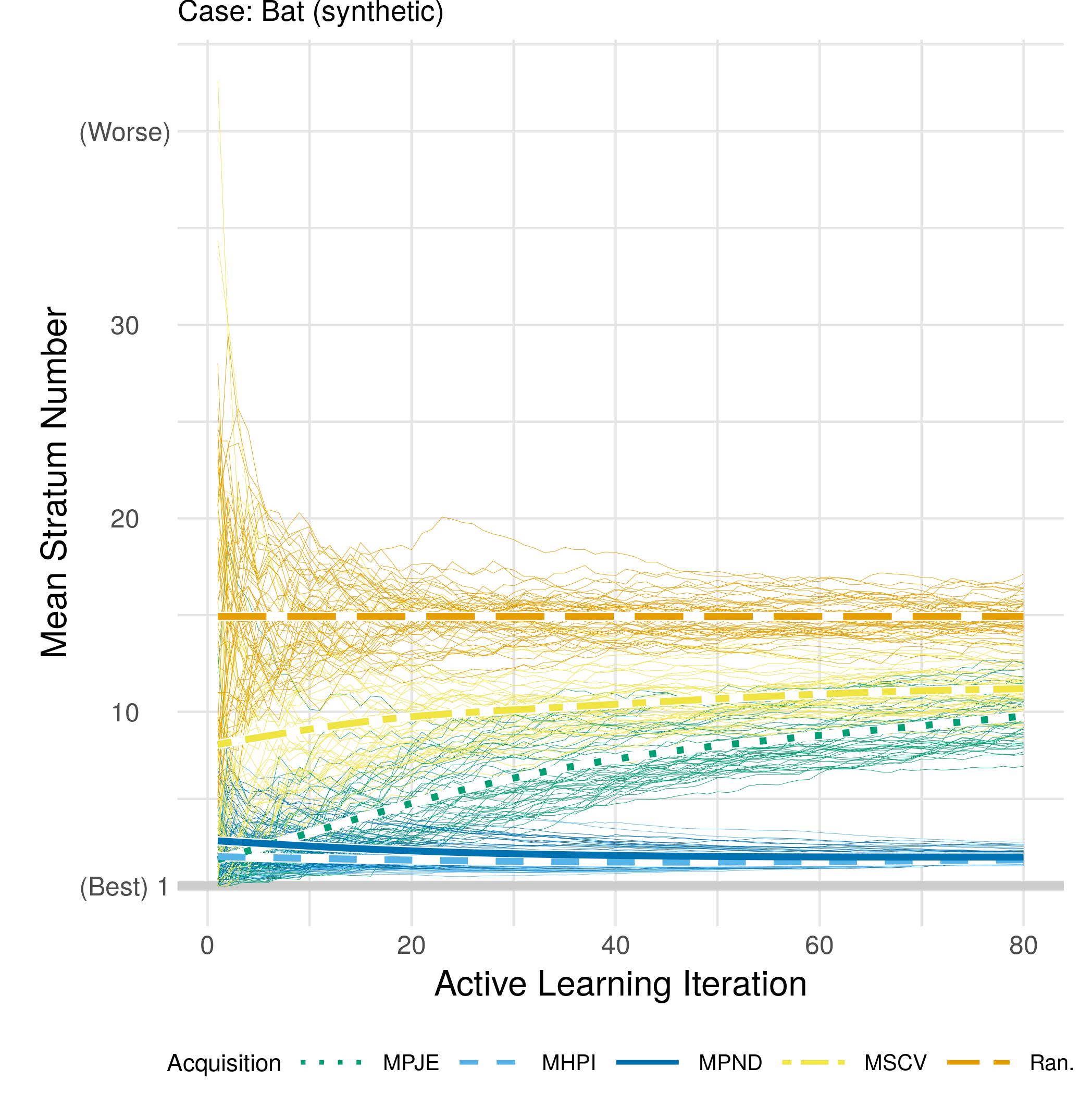}
  \end{minipage} %
  \caption{\emph{Mean stratum number highlights trends not captured by the number of non-dominated points (NNDP).} Results shown for the \texttt{Linear} (top-left), \texttt{Circular} (top-right), \texttt{Hyperbolic} (bottom-left) and \texttt{Bat} (bottom-right) synthetic test cases. Computing the mean stratum number reveals that the MHPI criterion acquires more near-frontier points than MPND in the early stages of AL, with a slight edge in the \texttt{Bat} case, and a decisive advantage in the \texttt{Hyperbolic} case. In the later stages of AL, MPND tends to dominate.}
  \label{fig:syn-mean-shell}
\end{figure}

Here we report results on candidate discovery performance. We consider both the usual metric of the number of non-dominated points (NNDP) acquired (Fig. \ref{fig:syn-total-nndp}) and the proposed measure of mean stratum number (Fig. \ref{fig:syn-mean-shell}). Across all test-cases considered, the acquisition functions that ultimately acquire the greatest NNDP are 1. MPND, 2. MHPI, and 3. MPJE---this matches descending order for Pareto frontier representation fidelity. Ranking results for the same criteria are mixed for shorter-term performance. These results suggest that capturing the true geometry of the Pareto frontier is most important in ``later stage'' AL, where the space of candidates is limited. As seen in Figure \ref{fig:motivating-example}, MPND also out-performs other criteria on the thermoelectric dataset in terms of long-run performance. These results suggest that incorporating high-fidelity modeling of the Pareto frontier to acquisition function is particularly important in well-studied problems with few possible candidates. Finally, note that the Hyperbolic test-case has only $5$ non-dominated candidates total---this fundamentally limits the resolution of AL results for this case. Figure \ref{fig:syn-mean-shell} analyzes the same AL results in terms of mean stratum number, which reveals other trends in the same data.

The difference in ``early stage'' performance is better revealed by the mean stratum results (Fig. \ref{fig:syn-mean-shell}). In the early stages of AL MHPI tends to give the best performance, delivering more candidates at or near the Pareto frontier, while MPND consistently achieves long-run performance. Note that MPJE performance tends to saturate earlier than other criteria, rather than ranking second when performance is measured with the NNDP metric---this suggests MPJE begins selecting candidates far from the Pareto frontier in later-stage AL. The reasons for these differences are explained in Subsection \ref{subsec:qualitative} below, which analyzes qualitative performance.

\clearpage
\subsection{Model improvement} \label{subsec:model}

\begin{figure}
  \centering
  \begin{minipage}{0.45\textwidth}
    \includegraphics[width=0.95\textwidth]{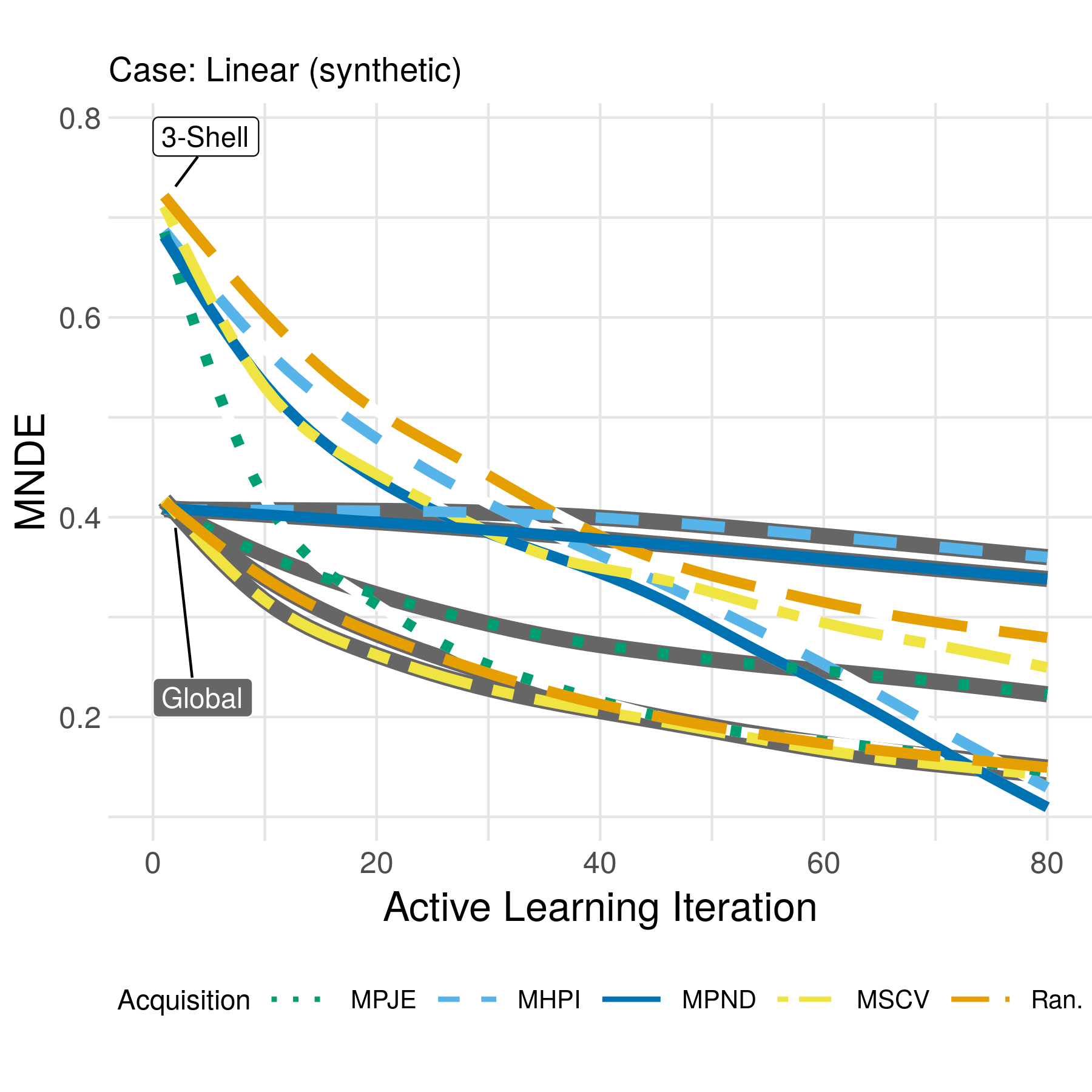}
  \end{minipage} %
  \begin{minipage}{0.45\textwidth}
    \includegraphics[width=0.95\textwidth]{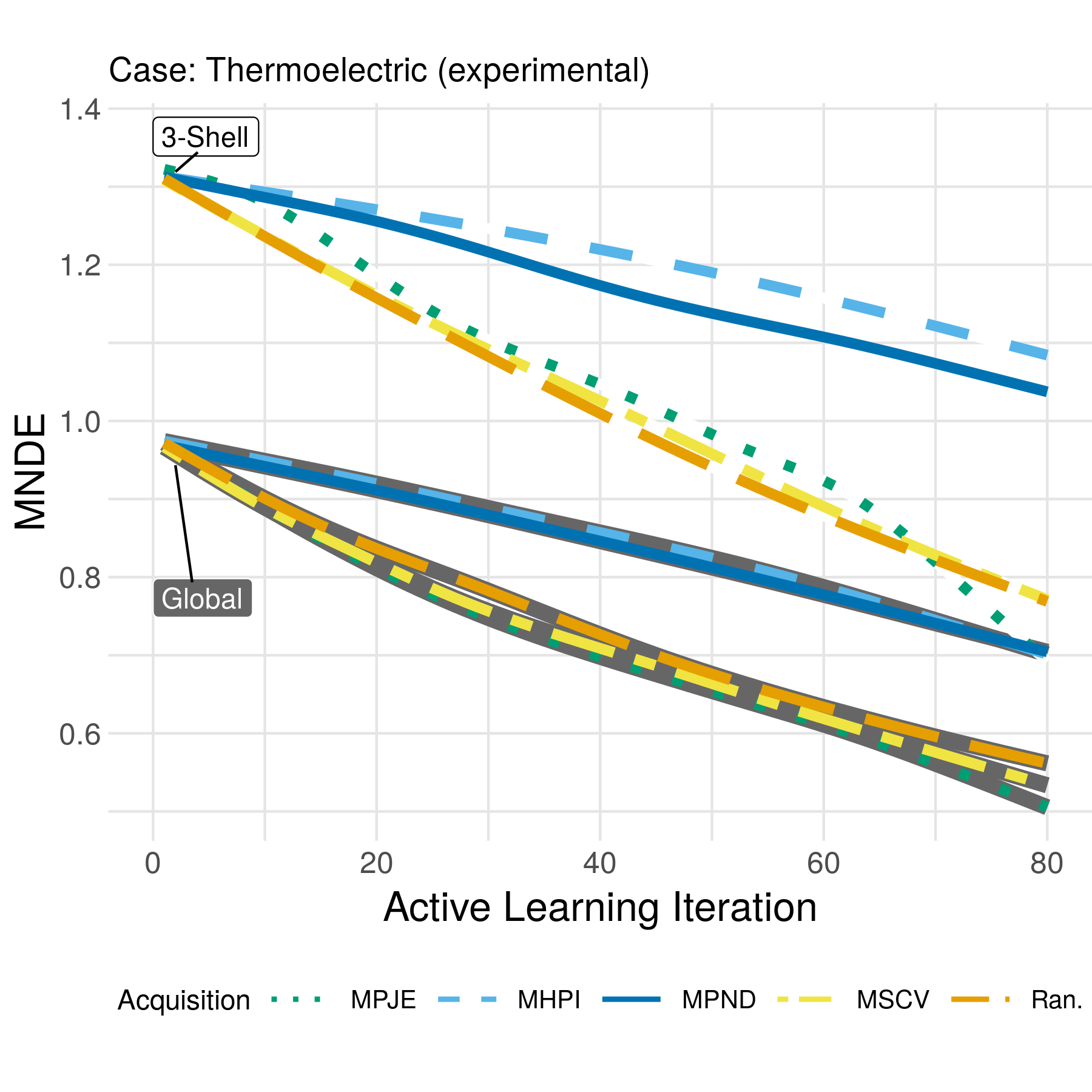}
  \end{minipage} %
  \caption{\emph{Best model accuracy improvement through active learning depends on error scope as well as acquisition function.} Mean non-dimensional error (MNDE) at different scopes (dark highlight for global, no highlight for shell) for the \texttt{Linear} (left) and thermoelectric (right) cases. The MSCV and Random selection criteria tend to result in strong error reduction in terms of global-scope across most test-cases. However, MPND tends to give the most consistent long-run performance in shell-scope for the synthetic cases. Note that in the \texttt{Linear} test-case, MPJE tends to achieve early gains for shell error, but saturates in performance as the acquisition function fails to spread across the Pareto frontier (Fig. \ref{fig:frontier-history}). The thermoelectric dataset does not exhibit these trends, showing similar error behavior in both scopes.}
  \label{fig:syn-error-all}
\end{figure}

Here we report results on how model accuracy evolves during AL. Figure \ref{fig:syn-error-all} summarizes consistent trends across all synthetic test cases considered: The Random and MSCV (uncertainty sampling) criteria tend to best improve model error in a global-scope, while MPND, MHPI, and MPJE tend to best improve error in the shell-scope. Generally MPJE gives a short-term reduction in shell error but reaches a plateau. The MPND and MHPI criteria give better long-term shell error reduction. Note that these trends do not persist for the thermoelectric dataset; based on the absolute magnitude of MNDE values for this case, we can tell that the model generally represents the data crudely. On this experimental dataset, the error reduction trends are similar across global- and shell-scopes.

\subsection{Qualitative performance} \label{subsec:qualitative}

\begin{figure}
  \centering
  \includegraphics[width=0.95\textwidth]{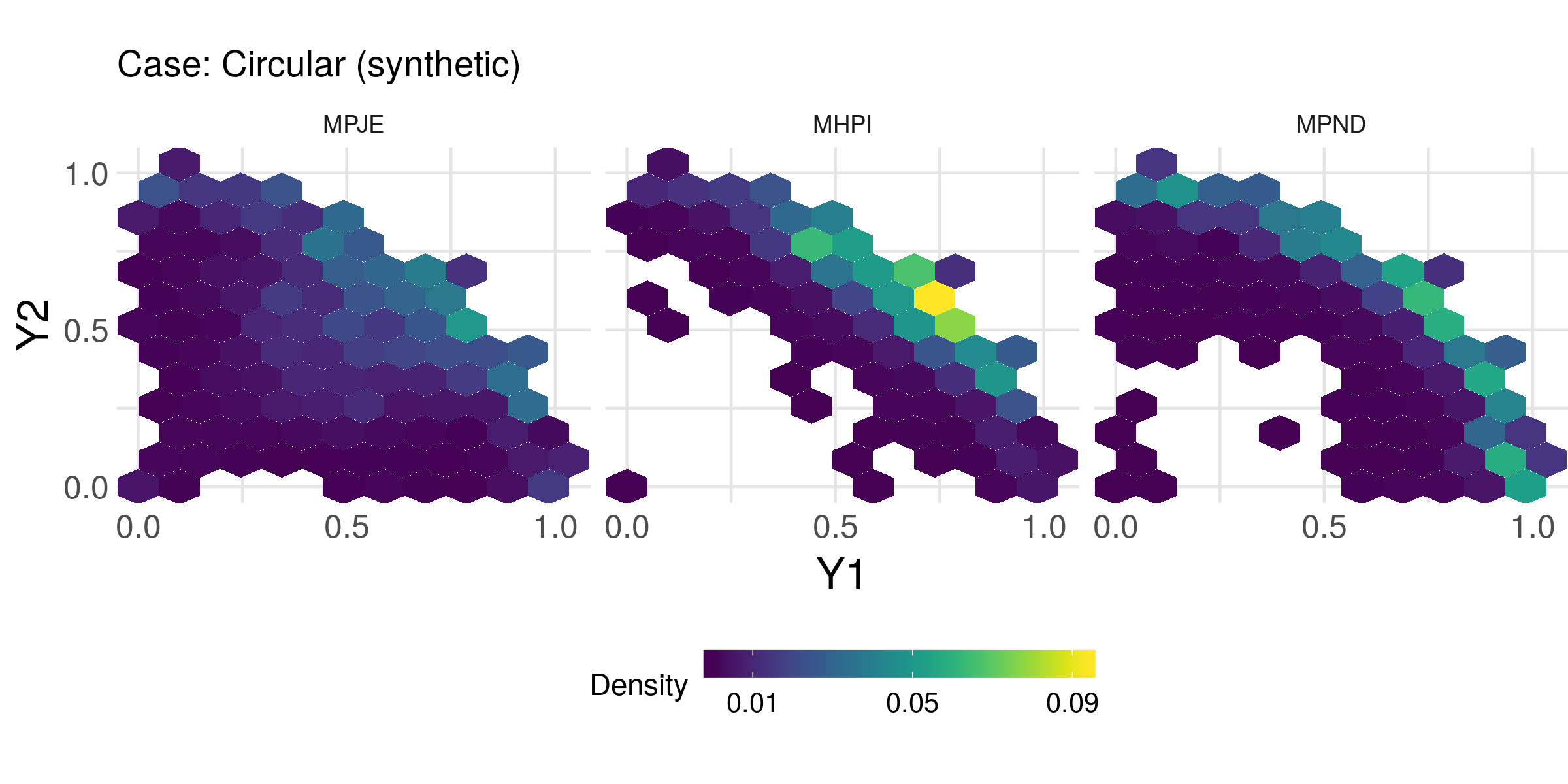}
  \includegraphics[width=0.95\textwidth]{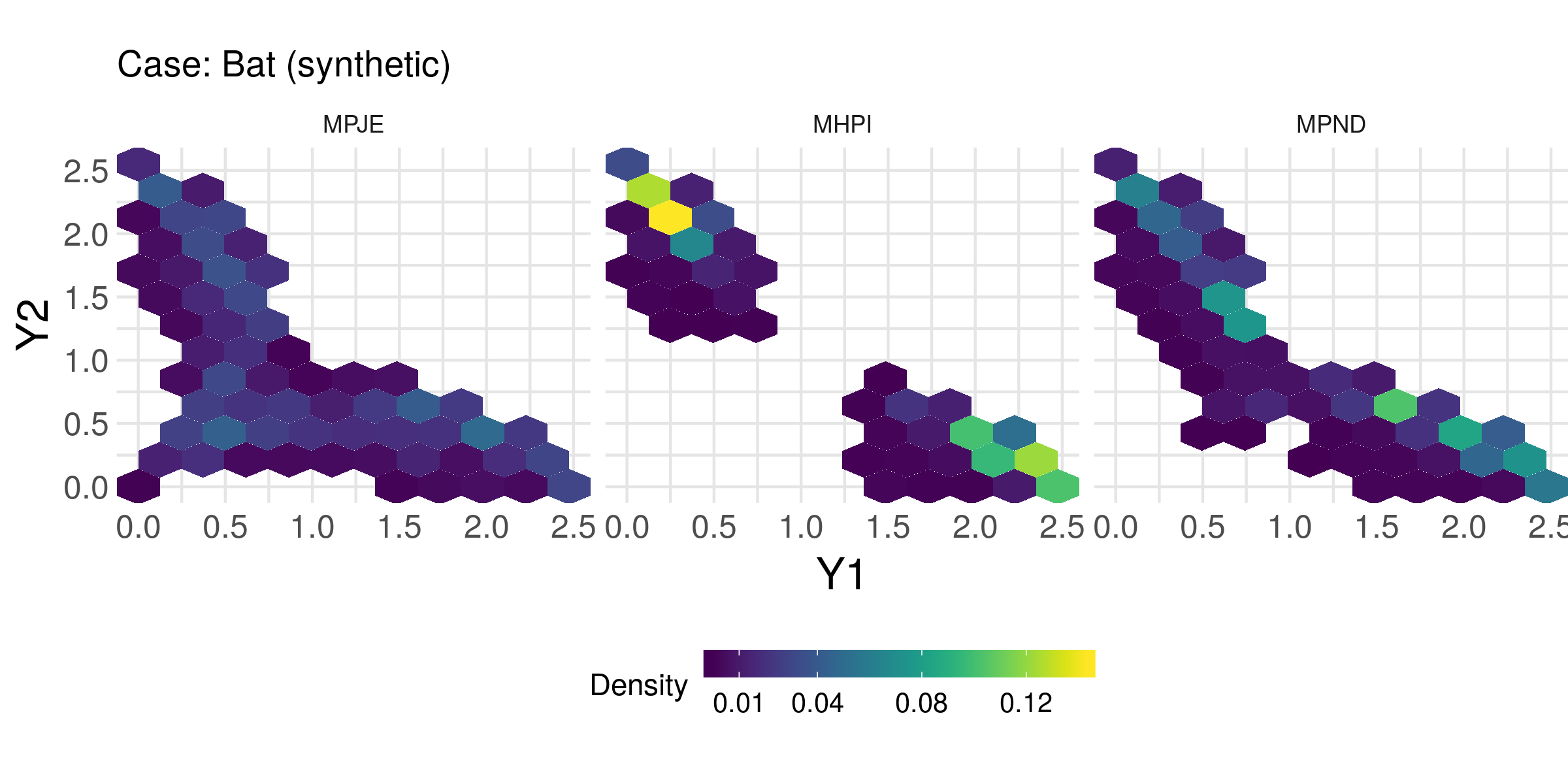}
  \caption{\emph{Higher-fidelity acquisition functions explore the frontier more thoroughly.} Density of selected points across all AL runs on the \texttt{Circular} and \texttt{Bat} cases, faceted by acquisition functions. Note that the selections made according to MPJE are scattered among the entire domain, as compared with MHPI and MPND. MPND tends to distribute its selections along the entire Pareto frontier, as opposed to MHPI, which concentrates on ``hot spots'' within the frontier. Note also that MPND has wider support than MHPI, indicating that the former criteria tends to (occasionally) select more strongly dominated candidates than the later.}
  \label{fig:frontier-history}
\end{figure}

Here we report how the acquisition functions of Subsection \ref{subsec:acq-fcn} behave in terms of qualitative performance: where their selections tend to lie in output space. Figure \ref{fig:frontier-history} reports empirical densities for selected candidates in output space. Broadly, MPJE tends to select candidates throughout the output space, MHPI focuses on ``hotspots'' along the frontier, and MPND thoroughly explores the frontier but infrequently selects strongly-dominated candidates. This qualitative analysis helps to explain the difference in NNDP (Fig. \ref{fig:syn-total-nndp}) and mean stratum (Fig. \ref{fig:syn-mean-shell}) performance: MPND is able to explore the frontier thoroughly, selecting many non-dominated points. However, MPND is also more liable than MHPI to select strongly non-dominated candidates, leading to a higher mean shell number. Conversely, MHPI tends to concentrate its selections closer to a limited region of the Pareto frontier, leading to fewer possible non-dominated selections but a lower mean stratum.

\clearpage
\section{Conclusions} \label{sec:conclusion}

In this work, we explored the relationship between different aspects of machine learning accuracy and candidate acquisition in multi-objective materials discovery. We showed that AL schemes which optimize for the usual notion of model accuracy---global error---do not guarantee optimal candidate discovery. To ameliorate the situation we introduced Pareto shell error, which we found to be more closely associated with discovering improved candidates.

We also studied the relationship between the accuracy with which acquisition functions represent the Pareto frontier and AL performance. We demonstrated that dimensionally-inhomogeneous acquisition functions can lead to non-robust decision making, and so limited our attention to dimensionally-homogeneous acquisition functions. We found that the long-run discovery of non-dominated candidates was improved by modeling the Pareto frontier with greater fidelity. However, an acquisition function which rendered the Pareto frontier with lesser fidelity (MHPI) uncovered more candidates at-or-near the Pareto frontier in the early stages of AL, leading to a lower mean stratum number than the highest fidelity acquisition function (MPND). We found that these acquisition functions tended to select material candidates at varying locations along the Pareto frontier with different frequency, leading to preferred ``hot-spots'' in material property space.

These results have ramifications for materials scientists seeking to use AL for materials discovery. Since global error is not always predictive of optimal candidate discovery, an analyst should check both global and shell error when deciding whether a model is sufficiently accurate to be used to rank materials candidates. Furthermore, an analyst can use these insights prescriptively, choosing hyperparameters to optimize shell error rather than global error.

The selection of an appropriate acquisition function is highly dependent on the analyst's goals. Based on our results, accurately modeling the Pareto frontier in an acquisition function is not critical for early-stage AL; that is, if the problem has a very large set of uncharacterized material candidates. Accurately capturing the Pareto frontier in the acquisition function logic becomes important when the design space is more thoroughly explored. In the absence of more specific preference criteria than non-dominated, the maximum probability non-dominated (MPND) strategy is the most performant among the acquisition functions tested here.

There are a number of remaining questions related to the present topic. By the simulation nature of our experiments, we were able to evaluate the true Pareto shell error; in practice one must devise an estimation technique based on available data. Given the suitability of different acquisition functions for different phases of active learning, a \emph{homotopy} approach that weights different objectives $\lambda f_1 + (1-\lambda) f_2$ with $\lambda$ varying between iterations may be a useful framework. In this work we fit independent models to all output quantities: Prior work suggests that modeling improvements can be made by accurately capturing output dependence structure among output quantities \cite{fricker2013multivariate,bhsstv2007}. It would be interesting to study whether similar improvements can be found in AL performance. A conceptually different way to treat the multi-objective setting is to turn some objectives into constraints; it would be interesting to compare the constrained approach against a multi-objective acquisition function in terms of mean stratum number (c.f. Fig. \ref{fig:syn-mean-shell}) and acquisition densities (c.f. Fig. \ref{fig:frontier-history}). Similarly, it would be interesting to compare performance across AL strategies using different scalarizations of the multi-objective space; e.g. the thermoelectric figure of merit $zT$.

\subsection*{Data Availability}

The Supplementary Materials contain scripts to reproduce the synthetic datasets used in this work, as well as a script to access and featurize the thermoelectric dataset used herein.

\subsection*{Author Contributions}

Z.dR., Y.K., E.A. and J.L. designed the research. Z.dR. performed the research and analyzed the data. Z.dR. and M.R. wrote the paper. All the authors read the manuscript, commented on the contents and agreed with the publication of the results.

\subsection*{Additional Information}

\paragraph{Competing interests:} The Authors declare no Competing Financial or Non-Financial Interests.


\bibliography{bibtex_database}
\bibliographystyle{plain}

\appendix

\section{Test Case Details} \label{apx:synthetic}

The following are the underlying distributions and function used to generate the synthetic data cases. These test cases were designed to provide a variety of Pareto frontier geometries. For all four synthetic test cases, the output space is two dimensional $\mY\in\R{2}$, and both outputs are to be maximized. Code for reproducing these datasets is provided in the Supplementary Materials.

\noindent\textbf{\texttt{Linear} Test Case}:

A simple linear frontier geometry, generated by rotating (and stretching) a uniform distribution.

\begin{equation} \begin{aligned}
    \mX_i &\sim \mathcal{U}[0, 1]^2, \\
    \mY &= [X_{i,1} - X_{i,2}, X_{i,1} + X_{i,2}]^{\top}.
\end{aligned} \end{equation}

\noindent\textbf{\texttt{Circular} Test Case}:

A circular frontier geometry, generated via trigonometric functions.

\begin{equation} \begin{aligned}
    X_{i,1} &\sim \mathcal{U}[0, 1], \\
    X_{i,2} &\sim \mathcal{U}[0, \pi/2], \\
    \mY_i &= [X_{i,1} \cos(X_{i,2}), X_{i,1} \cos(X_{i,2})]^{\top}.
\end{aligned} \end{equation}

\noindent\textbf{\texttt{Hyperbolic} Test Case}:

The hyperbolic responses lead to a far greater density of points near the origin. This results in a ``sparse'' Pareto frontier, which often has very few non-dominated candidates.

\begin{equation} \begin{aligned}
    \mX_i &\sim \mathcal{U}[0, 10]^2, \\
    \mY_i &= [1/X_{i,1}, 1/X_{i,2}]^{\top}.
\end{aligned} \end{equation}

\noindent\textbf{\texttt{Bat} Test Case}:

A non-convex Pareto frontier generated by perturbing the radius of the Circular test-case.

\begin{equation} \begin{aligned}
    X_{i,1} &\sim \mathcal{U}[0, 1], \\
    X_{i,2} &\sim \mathcal{U}[0, \pi/2], \\
    \mY_i &= [
      (X_{i,1} + 2|X_{i,2} - \pi/4|)\cos(X_{i,2}),
      (X_{i,1} + 2|X_{i,2} - \pi/4|)\sin(X_{i,2})
    ]^{\top}.
\end{aligned} \end{equation}

\end{document}

%% file: zachs_macros.tex
\usepackage{amsmath}  
\usepackage{amsfonts} 
\usepackage{scalerel,stackengine} 
\usepackage{graphicx} 
\usepackage{xcolor}   
\usepackage{caption}  
\usepackage{mathtools}
\usepackage{forest}   

\def\R#1{\mathbb{R}^{#1}}

\def\N#1{\mathbb{N}^{#1}}

\newcommand{\vsym}[1]{\boldsymbol{#1}}
\def\v#1{\vsym{#1}} 

\newcommand{\vw}{\boldsymbol{w}}
\newcommand{\vx}{\boldsymbol{x}}
\newcommand{\vy}{\boldsymbol{y}}

\newcommand{\msym}[1]{\boldsymbol{#1}}
\def\m#1{\msym{#1}} 

\newcommand{\mX}{\boldsymbol{X}}
\newcommand{\mY}{\boldsymbol{Y}}

\newcommand{\cA}{\mathcal{A}}

\newcommand{\cD}{\mathcal{D}}

\newcommand{\cM}{\mathcal{M}}

\newcommand{\cP}{\mathcal{P}}

\newcommand{\cS}{\mathcal{S}}

\newcommand{\cX}{\mathcal{X}}
\newcommand{\cY}{\mathcal{Y}}

\def\P{\mathbb{P}} 

\makeatletter
\def\mathcolor#1#{\@mathcolor{#1}}
\def\@mathcolor#1#2#3{%
  \protect\leavevmode
  \begingroup
    \color#1{#2}#3%
  \endgroup
}
\makeatother

\DeclareMathOperator*{\argmax}{arg\,max}

\def\i1{\textbf{1}}

\newcommand\blfootnote[1]{%
  \begingroup
  \renewcommand\thefootnote{}\footnote{#1}%
  \addtocounter{footnote}{-1}%
  \endgroup
}

\stackMath
\newcommand\reallywidehat[1]{%
\savestack{\tmpbox}{\stretchto{%
  \scaleto{%
    \scalerel*[\widthof{\ensuremath{#1}}]{\kern-.6pt\bigwedge\kern-.6pt}%
    {\rule[-\textheight/2]{1ex}{\textheight}}
  }{\textheight}%
}{0.5ex}}%
\stackon[1pt]{#1}{\tmpbox}%
}

\newcommand{\ftree}[1]{%
\begin{forest}
for tree={
    font=\ttfamily,
    grow'=0,
    child anchor=west,
    parent anchor=south,
    anchor=west,
    calign=first,
    edge path={
      \noexpand\path [draw, \forestoption{edge}]
      (!u.south west) +(7.5pt,0) |- node[fill,inner sep=1.25pt] {} (.child anchor)\forestoption{edge label};
    },
    before typesetting nodes={
      if n=1
        {insert before={[,phantom]}}
        {}
    },
    fit=band,
    before computing xy={l=15pt},
  }%
  #1%
\end{forest}
}

\usepackage{numdef}   
\num
\num
\num
\num
\num
\num
\num
\num

\num
\num
\num
\num
\num
\num
\num
\num

\num
\num
\num
\num
\num
\num
\num
\num

\num
\num
\num
\num
\num
\num
\num
\num

\num
\num
\num
\num
\num
\num
\num
\num

\num
\num
\num
\num
\num
\num
\num
\num

\num
\num
\num
\num
\num
\num
\num
\num

\num
\num
\num
\num
\num
\num
\num
\num

\num
\num
\num
\num
\num
\num
\num
\num

\num
\num
\num
